\documentclass[a4paper,fleqn]{cas-dc}

\usepackage[authoryear]{natbib}

\def\tsc#1{\csdef{#1}{\textsc{\lowercase{#1}}\xspace}}
\tsc{WGM}
\tsc{QE}

\begin{document}
\let\WriteBookmarks\relax
\def\floatpagepagefraction{1}
\def\textpagefraction{.001}

\shorttitle{PlantC2USeg}
\shortauthors{Yu Tian et~al.}

\title [mode = title]{PlantC2USeg: Cross-Scale Consistent Pre-Training for Few-Shot Unified Plant Point Cloud Segmentation}

\author[1]{Yu Tian}[orcid=0000-0001-9253-1914]
\ead{yu.tian2@mail.mcgill.ca}
\credit{Data curation, Investigation, Methodology, Visualization, Formal analysis, Writing - original draft, Writing - review \& editing}

\affiliation[1]{organization={McGill University},
            addressline={21111 Lakeshore Road},
            city={Sainte-Anne-de-Bellevue},
            postcode={H9X 3V9},
            state={Quebec},
            country={Canada}}

\author[1]{Xintong Jiang}[orcid=0000-0003-4947-1981]
\ead{xintong.jiang@mail.mcgill.ca}
\credit{Data curation, Investigation, Writing - review \& editing}

\author[1]{Jan Franklin Adamowski}[orcid=0000-0002-1242-3364]
\ead{jan.sdamowski@mcgill.ca}
\credit{Writing - review \& editing}

\author[1]{Shiv O. Prasher}
\ead{shiv.prasher@mcgill.ca}
\credit{Writing - review \& editing}

\author[1]{Shangpeng Sun}[orcid=0000-0001-7095-8626]
\ead{shangpeng.sun@mcgill.ca}
\credit{Conceptualization, Resources, Writing - review \& editing, Supervision, Project administration, Funding acquisition}
\cormark[1]
\cortext[cor1]{Corresponding author}

\begin{abstract}
As the advancement of modern crop breeding demands precise organ-level analysis for trait quantification, plant point cloud segmentation (PPCS) is elevated to a problem of vital importance.
Conventional deep learning approaches to PPCS rely heavily on densely annotated datasets, whose acquisition remains labor-intensive.
Human perception adapts rapidly from only a few novel examples, whereas unified PPCS adaptation from distribution-shifted examples with minimal additional training effort remains an open challenge.
Deep transfer learning (DTL) seeks to alleviate annotation dependency by pre-training on self-generated objectives, thereby structuring representation spaces that facilitate efficient adaptation to PPCS across diverse species, growth stages, and sensing conditions.
Inspired by the multi-scale verification process in manual annotation, we propose cross-scale consistency learning within the DTL framework, PlantC2USeg, to explicitly align features across spatial scales, together with an information-restricted decoding strategy that prevents reconstruction shortcuts and promotes robust adaptation.
Such pre-training enables stable few-shot generalization under species and sensor variations, while unified fine-tuning with inherited thresholds further reduces practical adaptation overhead.
Under full supervision on the Soybean3D dataset, the complete pre-training configuration improves semantic IoU and instance mWCov over random initialization, attaining the highest values among the compared methods at 91.91\% and 94.62\%, respectively.
With 20 labeled Soybean3D samples, PlantC2USeg leads both semantic IoU and instance mWCov at 89.78\% and 90.27\%, and with only 10 samples it retains the highest mWCov of 83.23\% while achieving 83.19\% IoU.
Across HR3D, 10-shot transfer to tobacco, tomato, and sorghum averages 78.41\% IoU and 79.42\% mWCov, while 22-shot transfer to SYAU-Maize attains the highest IoU and mRec among the compared methods at 92.75\% and 93.51\%.
Furthermore, achieving the leading category-averaged mIoU of 85.0\% on the ShapeNet Part benchmark demonstrates the general capability of the proposed framework to handle diverse shape variations beyond agricultural domains.
These results demonstrate that the proposed framework reduces the overall adaptation effort under distribution shifts, thereby enabling scalable plant phenotyping and transferable 3D representation learning to extend beyond agriculture.
\end{abstract}


\begin{keywords}
Point cloud segmentation \sep Deep transfer learning \sep Few-shot segmentation \sep Plant phenotyping
\end{keywords}

\maketitle

\section{Introduction}
High-throughput phenotyping (HTP) has enabled the acceleration of high-yield and stress-tolerant cultivar development in plant breeding \citep{pp, AIcrop}, which plays a vital role in addressing global food insecurity \citep{globalFoodDemand2021}.
By leveraging optical imaging technologies, HTP can select the ideal aboveground architecture by conducting rapid and non-destructive measurements of plant traits such as plant height and leaf angle \citep{yangHTP2020}.
Among the available sensing modalities, 3D plant point clouds have become a dominant representation for accurate phenotyping, as they capture details that are often occluded in 2D images.
Extracting traits from these data requires reliable segmentation that decomposes raw 3D point clouds into distinct plant organs.

Deep learning (DL) has become the dominant approach for plant point cloud segmentation (PPCS), owing to its ability to learn representations directly from raw 3D data and achieve superior performance compared with traditional handcrafted-feature pipelines \citep{Johnson1999Spin,Rusu2009FPFH}.
PointNet \citep{qi2017pointnet} made this shift possible for irregular point sets by directly consuming unordered points, avoiding conversion to voxel grids or rendered views.
By preserving permutation invariance through symmetric aggregation, PointNet leaves local metric neighborhoods largely implicit, which makes PointNet++ \citep{qi2017pointnet++} and DGCNN \citep{wang2019dgcnn} a natural extension for PPCS because their metric space grouping and dynamic $k$-NN graph construction strengthen local spatial cues for separating adjacent and morphologically similar plant organs \citep{heiwolt2021PointNet++inPP}.
More recently, PPCS has adopted Transformer \citep{vaswani2017transformer} architectures to introduce attention-driven contextual modeling for complex plant architectures, with PST \citep{du2023pst} targeting dense high-resolution rapeseed point clouds in which tiny siliques are spatially scattered and overlapping.
Despite these advances, supervised PPCS still depends on labeled plant point clouds, and performance on unseen cases remains governed by sample and distribution generalization \citep{rohlfs2025generalization}.
Such dependence on data scale becomes a fundamental bottleneck in practice, as acquiring dense annotations for plant point clouds is labor-intensive.

This limitation has motivated an increasing focus on developing models that remain stable under limited-annotation regimes, which is conceptually aligned with human intelligence that can learn from a few examples and adapt the knowledge to process new test cases \citep{george2017CAPTCHAs}.
Data-centric strategies have been explored to enrich the morphological variety, either by incorporating physical deformation on real-world labeled data \citep{yang2024deformation3d} or by generating synthetic plant point clouds through procedural modeling paradigms \citep{lee2023lSystem, du2025l1Gen}.
Although synthetic data provide scalable supervision, they are typically constructed under predefined morphological assumptions and simulation rules, the learned representations may therefore remain biased toward specific structural patterns.
Sparse-label propagation has also been investigated through AR-assisted annotation, where graph-based transductive inference derives pseudo point-level labels from limited manual inputs and reduces manual labeling to 32.3\% per sample \citep{li2025arplant}.
Practical HTP applications require models to generalize across point clouds collected from different species, growth stages, sensors, imaging settings, and environmental conditions.
This expectation exceeds the distributional scope of the data synthesis \citep{yang2024deformation3d, lee2023lSystem, du2025l1Gen} and label inference \citep{li2025arplant} strategies, positioning deep transfer learning (DTL) as a more scalable alternative by leveraging pre-trained representations to facilitate adaptation under distributional shifts \citep{sohail2025dtl}.

DTL frameworks address distributional shifts by decoupling representation learning from task-specific adaptation.
By optimizing self-generated tasks on unlabeled point clouds, unsupervised pre-training encourages the models to internalize structural regularities and morphological patterns that govern plant form beyond specific annotated datasets.
During downstream PPCS adaptation, segmentation fine-tuning constrains task-specific optimization within a representation space shaped by plant architectural patterns learned during pre-training.
Initializing segmentation models from these pre-trained representations mitigates distribution-specific overfitting under full supervision, while enabling robust feature reuse in limited-supervision settings and enhancing generalization across distributional shifts.
The growing affordability and portability of sensing systems, together with advances in reconstruction technologies, have improved the feasibility of large-scale plant point cloud collection \citep{harandi2023sensorAccessibility}, thereby strengthening the potential of DTL on abundant raw 3D phenotypic data.

Point cloud pre-training has explored several ways of deriving supervision from raw 3D shape, including patch reassembly \citep{sauder2019jigsaw3D}, occlusion completion \citep{wang2021occo}, orientation prediction \citep{poursaeed2020orientation3D}, autoregressive generation \citep{chen2023pointGPT}, cross-modal teacher distillation \citep{dong2023ACT}, and contrastive-generative learning \citep{qi2023recon}.
Masked point cloud modeling provides a partial-observation objective by predicting hidden tokens from visible point patches \citep{he2022mae,yu2022pointbert,pang2022pointmae}, with later extensions introducing multi-scale hierarchy \citep{zhang2022pointm2ae}, feature enhancement \citep{Zha2024PointFEMAE}, center prediction \citep{zhang2024pointpcpmae}, and decoupled cross-view reconstruction \citep{Zhang_2025_PointPQAE}.
Within PPCS, Eff-3DPSeg \citep{luo2023eff3dpseg} incorporated a viewpoint bottleneck (VIB) objective \citep{tian2022vibus} in which randomly transformed versions of the same plant point cloud were encoded and optimized to maximize cross-view consistency and channel decorrelation.
Soybean-PCMAE \citep{tian2025soybeanpcmae} incorporates cross-view consistency into 3D masked modeling and selectively predicts high-dimensional features of masked regions that exhibit high instability under perturbations.
The resulting representations are therefore regularized through whole-plant cross-view consistency or neighborhood reconstruction, with limited explicit coupling between fine-scale organ cues and broader plant architecture, which may constrain robustness under morphological variation.
The information constraint imposed by masked modeling can also be diluted during decoding, as current PPCS masked autoencoding frameworks \citep{xie2025plantmae,tian2025soybeanpcmae} process visible and masked tokens together through unrestricted self-attention.
From a representation learning perspective, masked modeling is fundamentally designed as an asymmetric generative task, in which masked regions are inferred solely from observable inputs.
By blurring this conditional structure, unrestricted decoding may weaken the encoder-centric regularization effect of the objective.

The effect of pre-training is ultimately realized during downstream PPCS adaptation, which primarily encompasses stem--leaf segmentation and leaf instance segmentation.
Stem--leaf semantic segmentation assigns organ-level labels to points, whereas leaf instance segmentation represents a more challenging task that further distinguishes spatially adjacent and morphologically similar leaves by assigning unique instance identifiers \citep{song2025PPCSReview}.
PST \citep{du2023pst} adopts a two-stage pipeline that first predicts semantic labels using a transformer-based backbone and subsequently performs instance-aware feature grouping to separate individual leaves.
Among general-purpose joint methods, SGPN \citep{wang2018sgpn} forms grouping proposals from pairwise similarities, whereas ASIS \citep{wang2019asis} uses semantic predictions to guide instance embeddings and aggregated instance features to refine semantics.
PlantNet \citep{li2022plantnet} proposes a dual-function segmentation network that simultaneously performs semantic classification and instance embedding, enabling leaf separation through MeanShift clustering \citep{Comaniciu2002meanshift} in the learned feature space.
The follow-up work, PSegNet \citep{li2022psegnet}, extends the dual-function architecture by introducing voxelized farthest point sampling and dual-scale feature extraction, promoting hierarchical encoding of plant geometric structures.
The reliance on mean-shift clustering within the learned embedding space introduces sensitivity to bandwidth selection, such that robust instance segmentation performance is contingent upon dataset-specific parameter tuning.
Deformation3D \citep{yang2024deformation3d} applies physically based deformation that preserves local organ geometry before training separate PointNet++ semantic and HAIS instance models.
Under the DTL paradigm, Eff-3DPSeg \citep{luo2023eff3dpseg} fine-tunes the pre-trained backbone using two independent networks, as unified optimization cannot be achieved stably.
One network is dedicated to stem--leaf semantic segmentation, while a separate instance segmentation network introduces an additional offset branch to predict point-wise displacements toward leaf centroids, followed by dual-set grouping on original and shifted coordinates following the PointGroup \citep{jiang2020pointgroup} strategy.
Soybean-PCMAE \citep{tian2025soybeanpcmae} adopts a two-stage fine-tuning strategy, first optimizing stem--leaf semantic segmentation and then extending the model with centroid-directed offset regression and instance-level discrimination supervision for leaf instance segmentation. During inference, leaf instances are obtained via clustering based on the predicted offsets.

\begin{figure*}[pos=htbp]
\centering
\includegraphics{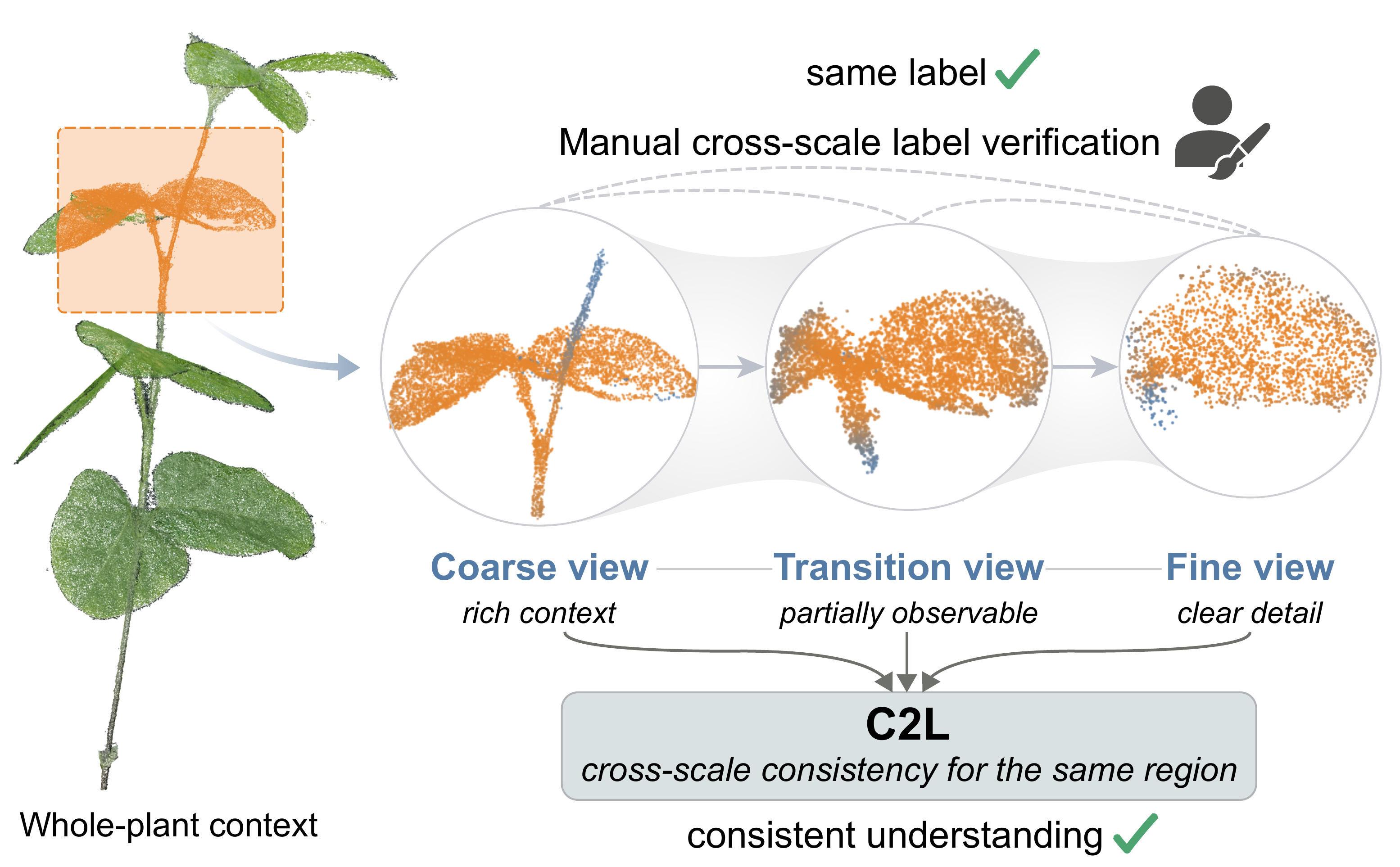}
\caption{Annotators routinely cross-check the same plant region across scales to resolve local ambiguity and confirm semantic identity. Inspired by this process, our cross-scale consistency learning (C2L) aligns representations of matched regions across scales, yielding scale-consistent point cloud understanding.}\label{fig1}
\end{figure*}

Existing PPCS transfer studies therefore motivate a pre-training design that preserves plant evidence across scales and avoids weakening the representation before it is used for unified semantic and instance fine-tuning.
In practice, annotators repeatedly alternate between detailed inspection and global assessment, verifying whether fine-grained interpretations remain consistent with organ arrangement and whole-plant architecture (Figure~\ref{fig1}).
This observation motivates an explicit cross-scale consistency principle in pre-training, where representations are learned to preserve coherence between local cues and higher-level plant organization.
For masked modeling to support PPCS transfer, the information available during decoding should keep masked regions dependent on visible plant evidence shaped by cross-scale consistency learning.
From the downstream perspective, unified PPCS adaptation requires stem--leaf semantic segmentation and leaf instance segmentation to remain stable within the same learned representation.
Current DTL pipelines separate or stage semantic and instance optimization, whereas fully supervised unified models \citep{li2022plantnet, li2022psegnet} still rely on manually tuned clustering to delineate leaf instances.
These observations motivate a unified few-shot adaptation strategy that integrates semantic and instance objectives within a shared fine-tuning process while reducing dataset-specific threshold tuning during instance inference.
Hence, we develop a cross-scale consistency pre-training framework with information-restricted decoding for PPCS, enabling robust representation learning and stable unified fine-tuning of semantic and instance objectives under few-shot supervision.
Our main contributions are as follows.
\begin{enumerate}
    \renewcommand{\labelenumi}{(\roman{enumi})}
    \item We formulate cross-scale consistency as a pre-training objective that draws inspiration from the cross-scale validation inherent in manual annotation (Figure \ref{fig1}), while introducing an information-restricted decoder to mitigate reconstruction shortcuts.
    \item We enable joint optimization of stem--leaf semantic segmentation and instance regularization in feature and coordinate spaces.
    Instance inference is performed via progressive segmentation with thresholds directly inherited from the fine-tuning stage, eliminating dataset-specific parameter sweeping.
    \item Our framework achieves improved fully supervised accuracy and maintains stable performance under few-shot adaptation with as few as 10 to 20 labeled samples. We systematically evaluate generalization across distributional shifts, including single- and multi-species datasets as well as cross-sensor scenarios.
    \item We extended Soybean3D dataset with additional reconstructed samples, consisting of 314 samples, among which 145 are labeled. The point count per sample ranges from 200,\allowbreak000-point to 1,\allowbreak000,\allowbreak000-point level, providing a high-quality benchmark for organ-level segmentation.
\end{enumerate}

\section{Methodology}
\subsection{Overview}
\begin{figure*}[pos=htbp]
\centering
\includegraphics[width=\textwidth]{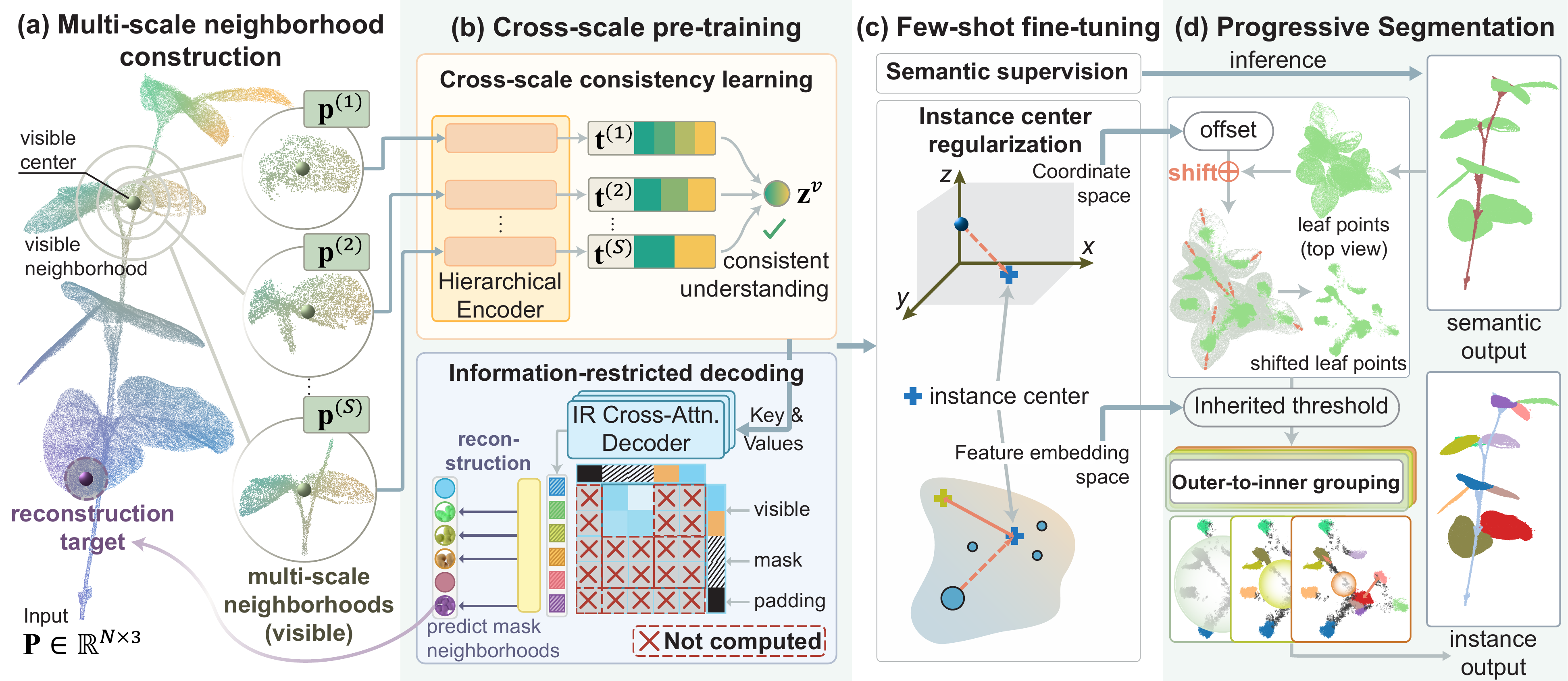}
\caption{PlantC2USeg provides an end-to-end pipeline for unified stem--leaf semantic segmentation and leaf instance segmentation on plant point clouds.
(a) Multi-scale neighborhood construction organizes the input point cloud $\mathbf{P}\in\mathbb{R}^{N\times 3}$ into a hierarchy of centers and their neighborhoods, with the visible subsets shown at each scale.
The purple target marks one schematic first-scale neighborhood withheld for reconstruction.
(b) A hierarchical encoder produces visible tokens across scales, and scale-specific projectors map them into a shared embedding space where cross-scale consistency relates representations of corresponding plant regions.
Information-restricted decoding reuses these projected visible representations as keys and values to condition masked queries that reconstruct first-scale targets, one of which is illustrated in panel (a).
(c) The few-shot setting illustrates unified fine-tuning, in which semantic supervision and complementary instance regularization in feature and coordinate spaces adapt the encoder initialized by cross-scale pre-training.
(d) Semantic output and predicted offsets enter progressive segmentation, which shifts leaf points toward estimated centers before grouping them into leaf instances.
Thresholds inherited from fine-tuning guide the grouping, and outer-to-inner processing is used for plants with broad leaves.}\label{fig2}
\end{figure*}

Figure~\ref{fig2} presents PlantC2USeg as combining cross-scale pre-training, unified fine-tuning, and progressive segmentation for stem--leaf semantic segmentation and leaf instance segmentation under limited annotation.

\subsection{Multi-scale Neighborhood Construction}
As illustrated in Figure~\ref{fig2}(a), the input point cloud $\mathbf{P}\in\mathbb{R}^{N\times 3}$ is organized into local neighborhoods that serve as contextual units for encoding, following Point-BERT and Point-M2AE \citep{yu2022pointbert,zhang2022pointm2ae}.
For scale $s\in\{1,\dots,S\}$, the center set is denoted as $\mathbf{P}_s\in\mathbb{R}^{N_s\times 3}$, with neighborhood indices $\mathcal{I}^{(s)}\in\mathbb{R}^{N_s\times k_s}$.
The neighbors are normalized with respect to their center point, so that each neighborhood is represented in a local reference frame that emphasizes relative spatial arrangement.
Finer-scale neighborhoods retain details around organ boundaries, whereas coarser-scale neighborhoods situate these details within a broader spatial extent.
This hierarchy provides local and broader neighborhood units over the plant point cloud, which are used in the subsequent cross-scale consistency objective.
\subsection{Cross-scale pre-training}
As shown in Figure~\ref{fig2}(a)--(b), cross-scale pre-training combines consistency among projected representations of corresponding visible plant regions with information-\linebreak restricted decoding that conditions reconstruction on those same representations.
The reconstruction signal recovers the center-relative input-point coordinate set around each masked first-scale center from visible plant evidence.
\subsubsection{Hierarchical feature extraction from visible neighborhoods}

Following Point-M2AE \citep{zhang2022pointm2ae}, the inherited hierarchical masking strategy samples one random mask at the coarsest scale and recursively propagates visibility to finer scales through the stored neighborhood indices.
At each scale $s$, the selected centers form the visible set $\mathbf{P}_s^{v} \in \mathbb{R}^{N_s^{v} \times 3}$, while the complement forms the masked set $\mathbf{P}_s^{m} \in \mathbb{R}^{N_s^{m} \times 3}$, with corresponding neighborhood indices $\mathcal{I}^{(s),v}$ and $\mathcal{I}^{(s),m}$.
Only at the first scale do visible neighborhoods supply direct point input to the encoder as sets of input-point coordinates expressed relative to their centers, while the corresponding sets around masked first-scale centers are withheld as reconstruction targets.
The first encoder stage maps the visible first-scale neighborhoods to $\mathbf{T}_1^{v}$, and each later stage aggregates groups of visible child tokens indexed by $\mathcal{I}^{(s),v}$ to produce $\{\mathbf{T}_s^{v}\}_{s=1}^{S}$.
These per-scale visible tokens serve as inputs to the scale-specific projectors used for cross-scale consistency learning.
\subsubsection{Cross-scale Consistency Learning}
\begin{figure*}[pos=htbp]
\centering
\includegraphics[width=\textwidth]{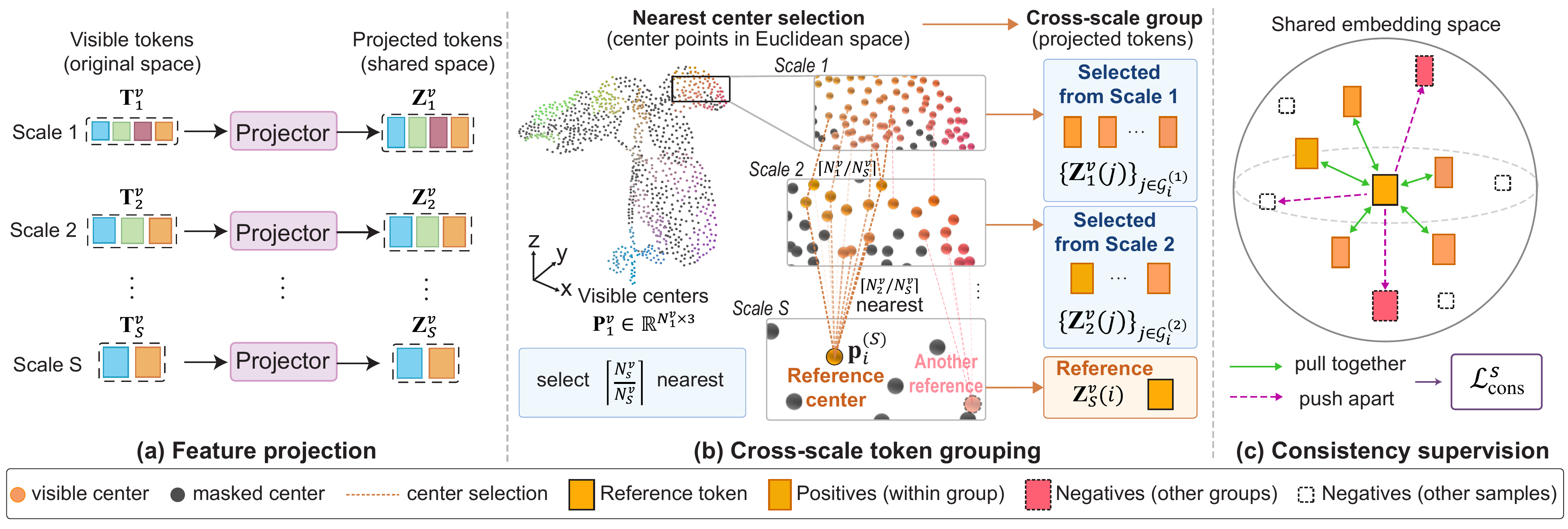}
\caption{Cross-scale consistency learning for visible plant regions.
(a) Visible tokens from each encoder scale are mapped by scale-specific projectors from their original feature spaces into a shared embedding space.
(b) In Euclidean space, each coarsest-scale reference center $\mathbf{p}_i^{(S)}$ selects the nearest finer-scale visible centers, with $\lceil N_s^{v}/N_S^{v}\rceil$ centers selected at scale $s$.
The selected indices form $\mathcal{G}_i^{(s)}$, and the corresponding projected tokens, such as $\{\mathbf{Z}_1^{v}(j)\}_{j\in\mathcal{G}_i^{(1)}}$ and $\{\mathbf{Z}_2^{v}(j)\}_{j\in\mathcal{G}_i^{(2)}}$, are grouped with the reference token $\mathbf{Z}_S^{v}(i)$.
(c) In the shared embedding space, the consistency loss pulls the reference token toward its within-group positives and pushes it away from negatives from other groups and other samples.}\label{fig3}
\end{figure*}

When examining a plant point cloud, human observers relate fine-scale organ details to organ arrangement and whole-plant architecture to assess whether the evidence remains coherent across scales.
To encode an analogous relation, cross-scale consistency learning maps visible tokens from each encoder stage into a shared embedding space, groups projected representations of corresponding plant regions through nearest-center associations, and regularizes their within-group consistency as shown in Figure~\ref{fig3}.

\paragraph{Feature projection}
Multi-scale token representations differ not only in spatial extent but also in their feature dimensionality. To make tokens from different representation levels directly comparable, a two-layer MLP projection network is applied at each level to map them into a shared embedding space. The projected features are $\ell_2$ normalized, yielding hyperspherical embeddings suitable for metric-based comparison. The projected representations are denoted as $\mathbf{Z}_s^{v} \in \mathbb{R}^{N_s^{v} \times D}$.
\paragraph{Cross-scale token grouping}
Projected representations at finer scales ($s<S$) are grouped around reference representations defined at the coarsest scale $S$ using rank-based nearest center selection \citep{cover1967nearest}.
For the $i$-th reference representation, the corresponding center point is denoted as $\mathbf{p}_i^{(S)} \in \mathbf{P}_S^{v}$.
At a finer scale $s<S$, candidate centers $\mathbf{p}_j^{(s)} \in \mathbf{P}_s^{v}$ are ranked by their Euclidean distance to $\mathbf{p}_i^{(S)}$.
The distance is computed as $d_{ij}^{(s)} = \left\| \mathbf{p}_i^{(S)} - \mathbf{p}_j^{(s)} \right\|_2$.
The top $\left\lceil N_s^{v}/N_S^{v} \right\rceil$ finer-scale centers are retained for each coarsest-scale reference, following nearest neighbor grouping practice in point cloud encoders \citep{wang2019dgcnn,yu2022pointbert,zhang2022pointm2ae}.
The selected indices form:
\begin{equation}
\mathcal{G}_i^{(s)}
=
\left\{
j \in \{1,\dots,N_s^{v}\}
\;\middle|\;
\operatorname{rank}\!\left(d_{ij}^{(s)}\right)
\le
\left\lceil \frac{N_s^{v}}{N_S^{v}} \right\rceil
\right\}
\end{equation}
where $\operatorname{rank}(d_{ij}^{(s)})$ denotes the ascending rank of $d_{ij}^{(s)}$.

Based on the index sets $\{\mathcal{G}_i^{(s)}\}_{s<S}$ constructed at all finer scales, a cross-scale group for the $i$-th reference representation is defined as
\begin{equation}
\left\{ \mathbf{Z}_S^{v}(i) \right\} \cup
\bigcup_{s<S}
\left\{ \mathbf{Z}_s^{v}(j) \mid j \in \mathcal{G}_i^{(s)} \right\}
\end{equation}
where projected representations at finer scales may be associated with multiple reference groups.
Each group thus links finer-scale projections to a coarsest-scale reference token, making the cross-scale association explicit for consistency supervision.
\paragraph{Consistency supervision}
Based on the reference-centered groups, a cross-scale consistency learning loss is introduced to encourage agreement among representations within each group. Directly enforcing absolute agreement among projections within each group is prone to degenerate solutions, where representations collapse to a constant vector. Instead, consistency at the $s$-th scale is imposed in a relative manner through a batch-wise N-pair objective \citep{sohn2016npair}:
\begin{equation}
\mathcal{L}_{\mathrm{cons}}^s
=
-\sum_{i\in\mathcal{B}_S}
\sum_{j \in \mathcal{G}_i^{(s)}}
\log
\frac{
\exp\!\left(\alpha\, \mathbf{Z}_s^{v}(j)^{\top} \mathbf{Z}_S^{v}(i)\right)
}{
\sum_{b\in\mathcal{B}_S}
\exp\!\left(\alpha\, \mathbf{Z}_s^{v}(j)^{\top} \mathbf{Z}_S^{v}(b)\right)
}
\end{equation}
where $\mathcal{B}_S$ indexes the reference groups available in the mini-batch, $\mathcal{G}_i^{(s)}$ contains the finer-scale projections associated with reference group $i$, and $\alpha=64$ rescales the normalized similarities, following prior hyperspherical metric-learning \citep{deng2019arcface} and point-cloud representation learning \citep{rao2023pointglr} practice. In implementation, the summed terms are averaged over the positive associations $\{(i,j)\mid i\in\mathcal{B}_S, j\in\mathcal{G}_i^{(s)}\}$, and the final consistency loss is obtained by averaging $\mathcal{L}_{\mathrm{cons}}^s$ over all finer scales $s<S$.

\subsubsection{Information-Restricted Decoding}
\begin{figure*}[pos=htbp]
\centering
\includegraphics[width=\textwidth]{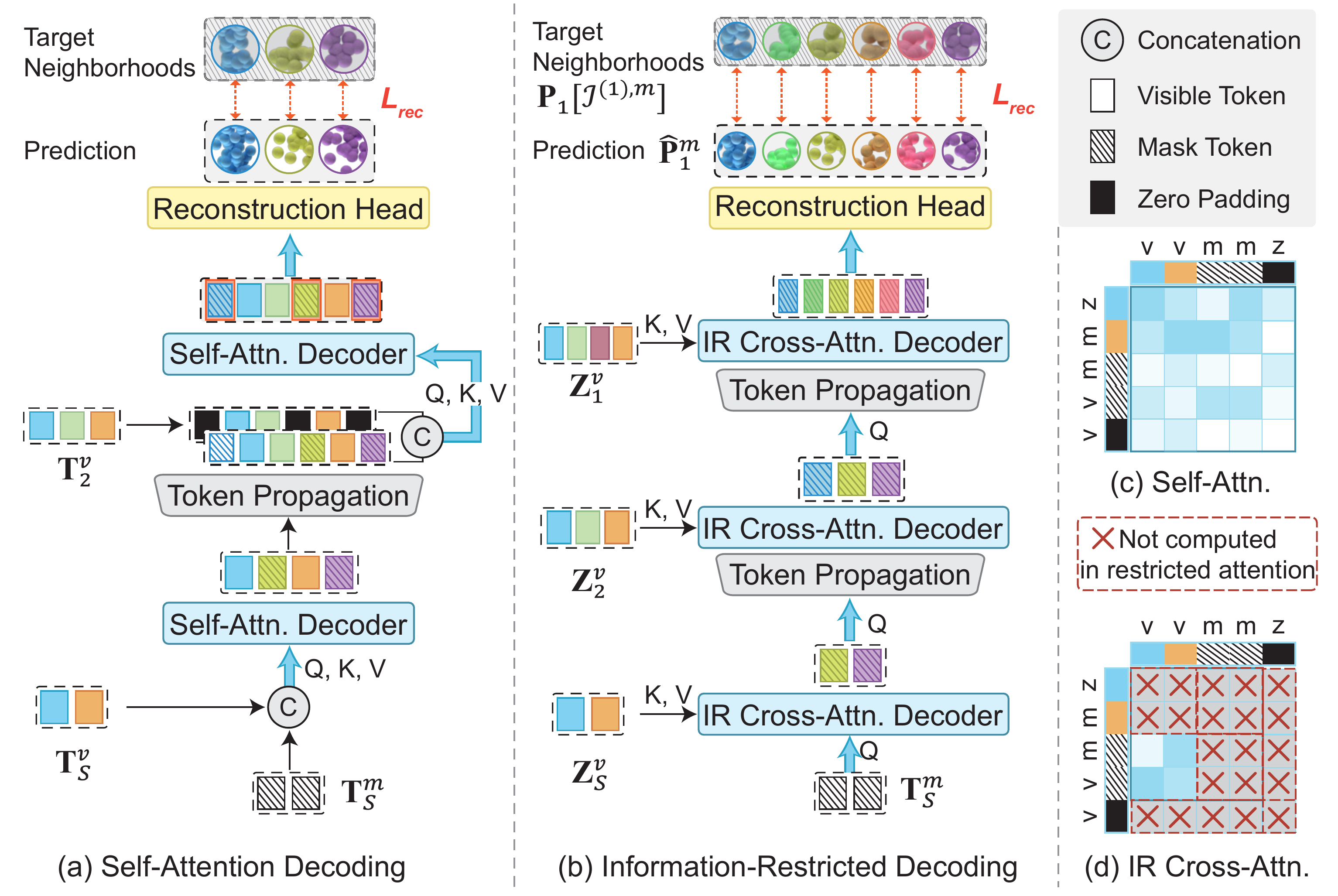}
\caption{Comparison between self-attention decoding and information-restricted decoding for hierarchical point cloud reconstruction.
(a) Self-attention decoding that freely mixes tokens across spatial regions.
(b) Information-restricted decoding that reconstructs masked neighborhoods by querying visible representations only.
(c) Self-attention incurs quadratic complexity due to all-to-all attention.
(d) Structured restricted attention map.}\label{fig4}
\end{figure*}

Cross-scale consistency constrains finer-scale representations to align with the coarsest projections, yet provides no direct supervision on the validity of the final-stage encoded features themselves. As a result, suboptimal coarse representations may consistently bias learning throughout the hierarchy.
Decoding is therefore used to anchor the encoded features through masked-neighborhood reconstruction, with information restrictions applied to reduce mutual interactions among masked tokens.
\paragraph{Self-Attention Decoding and its limitations in Point Clouds}
In self-attention decoders, reconstruction is performed through $S\!-\!1$ decoding stages operating from coarse to fine scales, as illustrated in Figure~\ref{fig4}(a). Decoding is initialized by combining the coarsest visible tokens $\mathbf{T}_S^{v}$ with learnable mask tokens $\mathbf{T}_S^{m} \in \mathbb{R}^{N_S^{m} \times D}$, and self-attention is applied to infer masked representations. At each stage, self-attention is performed globally over the concatenated token set, such that queries, keys, and values are jointly derived from both visible and masked token types. The updated tokens are propagated to finer scales following the feature propagation strategy in PointNet++ \citep{qi2017pointnet++}, after which they are concatenated with visible tokens from the corresponding encoder scale to serve as input for the next stage. The decoder output at the final $(S\!-\!1)$-th stage is used to reconstruct the masked local neighborhoods by applying a reconstruction head to the masked tokens, with supervision provided by the Chamfer Distance. In point clouds, geometry is encoded explicitly through positional embeddings. Unconstrained self-attention decoding permits masked tokens to exchange meaningful spatial cues within the attention mechanism, thereby allowing positional correlations to influence reconstruction more strongly than semantic cues from visible neighborhoods. To mitigate these unrestricted interactions, an information-restricted decoder based on cross-attention is introduced.
\paragraph{Restricted information flow}
The information-restricted formulation preserves the coarse-to-fine decoding structure while enforcing asymmetric information flow through cross-attention (Figure~\ref{fig4}(b)).
Masked tokens are initialized as learnable embeddings, with their role in decoding restricted to querying the projected visible representations from the consistency learning objective.
The cross-attention block design is inspired by the vanilla cross-attention decoder introduced in CrossMAE \citep{fu2024crossMAE}. This design is extended to hierarchical point cloud decoding, and an attention mask is applied as an additional constraint to filter out invalid interactions during decoding.
Batch-wise processing under multi-scale masking requires padding masked token embeddings at scales $s < S$ to the maximum masked length
$\widehat{N}_s^{m} = \max_b N_{s,b}^{m}$. In each block, attention is computed between padded masked tokens and the projected visible representations $\mathbf{Z}_s^{v}$ via an information-restricted cross-attention, computed as:
\begin{equation}
\begin{aligned}
&\mathrm{IR\text{-}CrossAttn}(\widetilde{\mathbf{T}}_s^{m}, \mathbf{Z}_s^{v})
\\
&=
\mathbf{M}_s\odot\left[
\mathrm{softmax}\!\left(
\frac{
(\widetilde{\mathbf{T}}_s^{m} \mathbf{W}_Q)
(\mathbf{Z}_s^{v} \mathbf{W}_K)^\top
}{\sqrt{d}}
\right)
(\mathbf{Z}_s^{v} \mathbf{W}_V)
\right]
\end{aligned}
\end{equation}
where $\widetilde{\mathbf{T}}_s^{m}$ denotes a padded embedding of the masked tokens $\mathbf{T}_s^{m}$,
$\mathbf{M}_s \in \{0,1\}^{\widehat{N}_s^{m}\times 1}$ equals one at valid masked-token positions and zero at padded positions,
and $d$ denotes the dimensionality of each attention head. The mask is broadcast over the feature dimension to suppress padded outputs.

As shown in Figure \ref{fig4}~(c), dense pairwise interactions among all tokens are induced by self-attention, leading to quadratic complexity with respect to the token count and redundant computation over padding and masked positions. In Figure \ref{fig4}~(d), attention is confined to masked-to-visible interactions, preventing information exchange among masked tokens.
This formulation decouples feature extraction from decoding, allowing reconstruction to be driven by geometrically grounded visible representations.
\paragraph{Reconstruction aligned with dense prediction tasks}
Dense prediction tasks such as segmentation rely on stable fine-scale spatial cues at early layers, motivating masked tokens to be updated by querying visible representations down to the finest scale.
The final masked-query representation is therefore denoted as $\widetilde{\mathbf{T}}_1^{m}$.
A single linear projection reconstructs the masked first-scale neighborhoods as $\widehat{\mathbf{P}}_1^m=\widetilde{\mathbf{T}}_1^m\mathbf{W}_{\mathrm{rec}}$, where $\mathbf{W}_{\mathrm{rec}}$ denotes the learnable projection of the reconstruction head.
Reconstruction is supervised using the $\ell_2$ Chamfer Distance \citep{fan2017l2cd}
\begin{equation}
\mathcal{L}_{\mathrm{rec}}
=
\sum_{\mathbf{p} \in \mathbf{P}_{1}[\mathcal{I}^{(1),m}]}
\min_{\widehat{\mathbf{q}} \in \widehat{\mathbf{P}}_1^m}
\|\mathbf{p} - \widehat{\mathbf{q}}\|_2^2
+
\sum_{\widehat{\mathbf{q}} \in \widehat{\mathbf{P}}_1^m}
\min_{\mathbf{p} \in \mathbf{P}_{1}[\mathcal{I}^{(1),m}]}
\|\widehat{\mathbf{q}} - \mathbf{p}\|_2^2
\end{equation}
\subsection{Unified few-shot fine-tuning}
Unified fine-tuning adapts the encoder initialized by cross-scale pre-training in one stage for stem--leaf semantic segmentation and leaf instance segmentation under full-shot and few-shot supervision.
At each encoder scale, representations learned for local neighborhoods are interpolated to the input points, concatenated with the coordinates of those points, and processed by a scale-specific MLP.
The three resulting point-level outputs are averaged to form the shared point-wise embedding $\mathbf{F}$.

\subsubsection{Semantic supervision}
\begin{figure*}[pos=htbp]
\centering
\includegraphics[width=\textwidth]{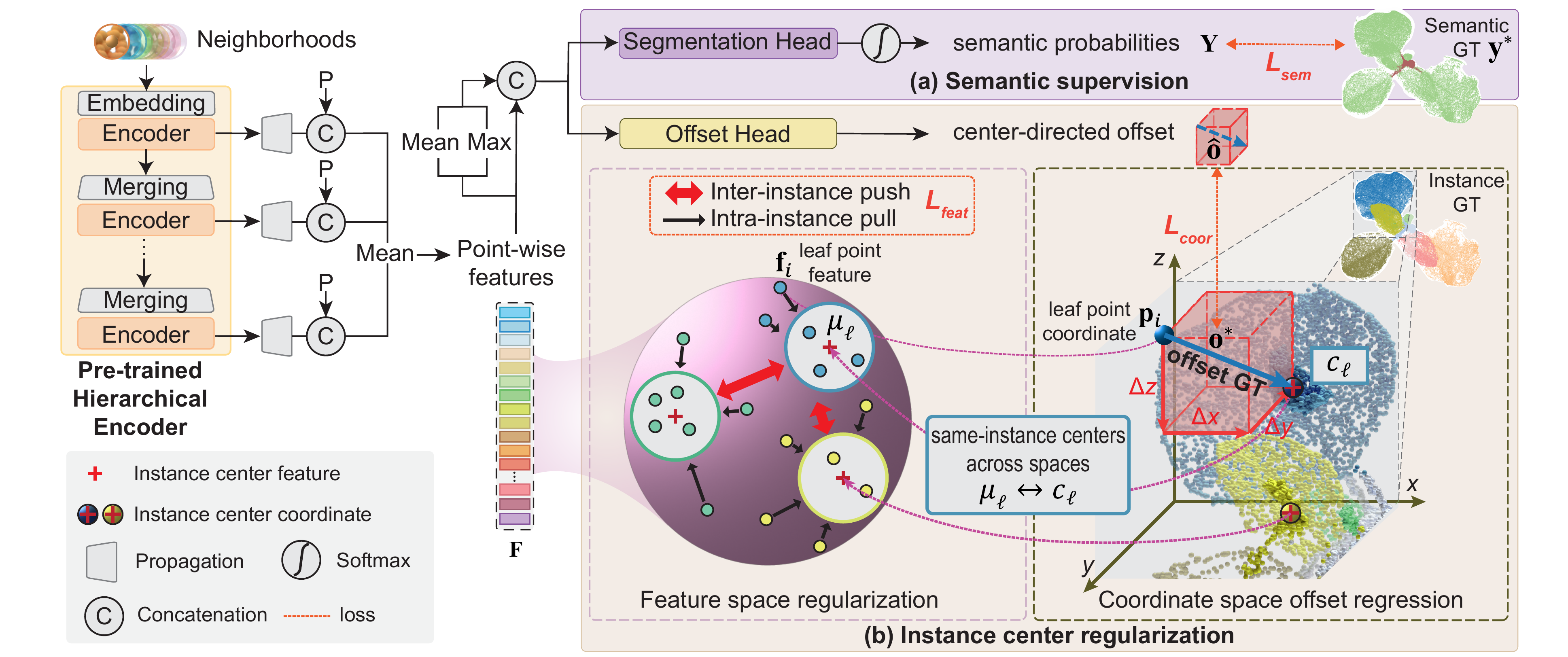}
\caption{Unified adaptation of the pre-trained hierarchical encoder.
Representations of local neighborhoods from the three encoder scales are propagated to the input points and averaged into $\mathbf{F}$.
Feature regularization acts directly on $\mathbf{F}$, while global mean and maximum summaries augment the point-level inputs to the semantic and offset heads.
(a) The semantic head predicts probabilities $\mathbf{Y}$, which are compared with ground-truth labels through $\mathcal{L}_{\mathrm{sem}}$.
(b) Under $\mathcal{L}_{\mathrm{feat}}$, embeddings of leaf points are drawn toward their leaf-specific feature centers $\boldsymbol{\mu}_\ell$, and centers of different leaves are separated.
The offset head predicts center-directed displacements from $\mathbf{p}_i$ to $\boldsymbol{c}_\ell$ under $\mathcal{L}_{\mathrm{coor}}$.}\label{fig5}
\end{figure*}

The semantic head supports stem--leaf semantic segmentation through a point-wise classification objective that updates the shared encoder.
For $N_{\mathrm{sem}}$ valid points, the head converts the globally augmented point representation into class probabilities $\mathbf{Y}_i$ (Figure~\ref{fig5}(a)), and the mean negative log-likelihood is
\begin{equation}
\mathcal{L}_{\mathrm{sem}}
=
-\frac{1}{N_{\mathrm{sem}}}\sum_{i=1}^{N_{\mathrm{sem}}}
\log Y_{i,y_i^{*}}
\end{equation}
where $y_i^{*}$ is the ground-truth semantic label and $Y_{i,y_i^{*}}$ is its predicted probability.

\subsubsection{Instance-center regularization}
For leaf instance segmentation, the coupled design constrains learned embeddings around leaf centers in feature space and supervises predicted offsets with displacements to the matching leaf centers in coordinate space.
Discriminative instance embedding \citep{de2017discriminativeloss} draws embeddings of leaf points toward their feature centers and separates centers that represent different leaves.
Center-directed offset regression \citep{jiang2020pointgroup,chen2021hais} uses the globally augmented point representation to predict a three-dimensional displacement from each leaf point to the center of that leaf in coordinate space (Figure~\ref{fig5}(b)).
Coordinate offsets provide positional guidance when spatially distinct leaves exhibit similar fine-scale shape cues, whereas proximity in feature space indicates instance affinity when leaf overlap makes center-directed offsets ambiguous.

\paragraph{Feature-space regularization}
Let $\mathbf{F} \in \mathbb{R}^{N_{\mathrm{feat}} \times F}$ contain the point-wise embeddings for the $N_{\mathrm{feat}}$ points evaluated by feature regularization.
For each leaf instance $\ell$, $\mathcal{A}_\ell$ contains the indices of its leaf points and excludes stem points.
The corresponding feature center is $\boldsymbol{\mu}_\ell = |\mathcal{A}_\ell|^{-1} \sum_{i \in \mathcal{A}_\ell} \mathbf{f}_i$.
For a sample with $L>1$, the discriminative instance embedding loss \citep{de2017discriminativeloss} formalizes within-leaf compactness, separation between centers of leaf instances, and center regularization as
\begin{equation}
\begin{aligned}
\mathcal{L}_{\mathrm{feat}}
=&
\frac{1}{L}
\sum_{\ell=1}^{L}
\frac{1}{|\mathcal{A}_\ell|}
\sum_{i \in \mathcal{A}_\ell}
\Big[
\lVert
\mathbf{f}_i - \boldsymbol{\mu}_\ell
\rVert_c
-
\delta_s
\Big]_+^{2}
\\
&+
\frac{1}{L(L-1)}
\sum_{\ell_a=1}^{L}
\sum_{\substack{\ell_b=1 \\ \ell_b \neq \ell_a}}^{L}
\Big[
2\delta_d
-
\lVert
\boldsymbol{\mu}_{\ell_a} - \boldsymbol{\mu}_{\ell_b}
\rVert_c
\Big]_+^{2}
\\
&+
\lambda_{\mathrm{ctr}}
\frac{1}{L}
\sum_{\ell=1}^{L}
\lVert
\boldsymbol{\mu}_\ell
\rVert_c
\end{aligned}
\end{equation}
where $[x]_+ = \max(0,x)$ is the hinge function and feature distances use $c=1$, following ASIS \citep{wang2019asis}.
The margin $\delta_s$ bounds the distance from each embedding to its feature center, $2\delta_d$ specifies the minimum distance between centers, and $\lambda_{\mathrm{ctr}}$ weights center regularization.
When $L=1$, the inter-instance term is omitted.

\paragraph{Coordinate-space offset regression}
For a leaf point $\mathbf{p}_i \in \mathbb{R}^{3}$ assigned to leaf instance $\ell$, the target offset is
\begin{equation}
\mathbf{o}_i^{\ast} = \boldsymbol{c}_\ell - \mathbf{p}_i
\end{equation}
where $\boldsymbol{c}_\ell = |\mathcal{A}_\ell|^{-1} \sum_{i \in \mathcal{A}_\ell} \mathbf{p}_i$ is the center of leaf instance $\ell$ in coordinate space.
For $N_{\mathrm{coor}}$ evaluated points, the offset head predicts $\hat{\mathbf{o}}_i \in \mathbb{R}^3$, and the masked L1 objective is
\begin{equation}
\mathcal{L}_{\mathrm{coor}}
=
\frac{
\sum_{i=1}^{N_{\mathrm{coor}}}
\mathbf{1}_i
\left\|
\hat{\mathbf{o}}_i - \mathbf{o}_i^{\ast}
\right\|_1
}{
\sum_{i=1}^{N_{\mathrm{coor}}}
\mathbf{1}_i + \epsilon
}
\end{equation}
where $\mathbf{1}_i=1$ for leaf points with $y_i^{*} \in \mathcal{Y}_{\mathrm{inst}}$ and $\mathbf{1}_i=0$ otherwise, and $\epsilon=10^{-6}$.
The three objectives jointly update the shared encoder through
\begin{equation}
\mathcal{L}
=
\mathcal{L}_{\mathrm{sem}}
+
\mathcal{L}_{\mathrm{feat}}
+
\mathcal{L}_{\mathrm{coor}}
\end{equation}
while aggregate feature-distance statistics from fine-tuning provide the three criteria for progressive segmentation.

\subsection{Progressive segmentation}
Progressive segmentation carries the complementary coordinate and feature cues learned during unified fine-tuning into the inference of individual leaf instances.
Semantic predictions select leaf points, and predicted offsets shift their coordinates toward estimated centers according to
\begin{equation}
\tilde{\mathbf{p}}_i
=
\mathbf{p}_i + \widehat{\mathbf{o}}_i
\qquad
\widehat{y}_i \in \mathcal{Y}_{\mathrm{inst}}
\end{equation}
where $\widehat{y}_i$ is the predicted semantic label of point $\mathbf{p}_i$.
Planar distances after coordinate shifting determine which points enter each radial step, whereas feature embeddings guide DBSCAN partitioning, reconnection across steps, assignment of residual points, and final consolidation.
Values for the three distance criteria are inherited from the statistics recorded during fine-tuning, avoiding manual threshold sweeps or recalibration on the labeled test set.
Distances from points to the feature center of their instance define $\epsilon_{\mathrm{db}}$.
The distance from each feature center to the nearest center of another instance defines $\epsilon_{\mathrm{rec}}$, while a high quantile of these within-instance distances defines $\epsilon_{\mathrm{merge}}$.

PlantC2USeg uses outer-to-inner progression for plants with broad leaves or rosette morphology.
Following radius-decremental grouping \citep{Roggiolani2023radiusdecremental}, this progression begins with peripheral points and advances toward the central stem region, where closely spaced leaf petioles make assignment more difficult.
For plants with narrow, grass-like leaves, radial progression is disabled and all candidates are processed in a single pass.
One planar center $\bar{\mathbf{p}}$ and the maximum radius are computed once from the shifted candidate set before radial processing.
At step $q$, let $\mathcal{U}^{(q)}$ denote the indices that remain unassigned at the start of the step, so the radially eligible set is
\begin{equation}
\mathcal{I}^{(q)} =
\left\{
i \in \mathcal{U}^{(q)} \;\middle|\;
\left\lVert
\tilde{\mathbf{p}}_{i,xy} - \bar{\mathbf{p}}_{xy}
\right\rVert_2 > r^{(q)}
\right\}
\end{equation}
where $r^{(q)}$ decreases linearly from the initial radius selected for the plant morphology and supervision regime to zero.

At each radial step, DBSCAN \citep{ester1996dbscan} partitions the eligible feature embeddings with the L1 radius $\epsilon_{\mathrm{db}}$, while noise and other unassigned points remain available to later inward steps.
The first successful partitions initialize the instance labels.
Each later partition is matched to the nearest existing instance by the L1 distance between their feature centroids.
It receives that instance label when the distance is below $\epsilon_{\mathrm{rec}}$ and receives a new label otherwise, after which the affected feature centroids are recomputed.
After the final radial step, residual points are assigned to the nearest existing feature centroid without a threshold and form one new group when no centroid exists.
Final consolidation completes the partition into leaf instances by merging groups whose feature centroids are separated by no more than $\epsilon_{\mathrm{merge}}$.
\section{Experiment}
\subsection{Datasets}
Plant species exhibit substantial variations in morphological structures, leading to distinct geometric characteristics in their corresponding point clouds. Experiments were conducted on a 3D soybean point cloud dataset originally introduced in \citep{luo2023eff3dpseg}, with additional samples used in this work, as well as two publicly available plant datasets, HR3D \citep{conn2017hr3d} and SYAU-Maize \citep{yang2024deformation3d}. In addition, ShapeNet Part \citep{yi2016shapeNetPart} was included as a complementary benchmark to assess the generality of the proposed method beyond plant-specific datasets, given the limited semantic diversity of existing plant point cloud datasets. The characteristics of each dataset are described below.
\paragraph{Soybean3D}
The soybean point cloud dataset used in this work consists of 314 point clouds reconstructed from multi-view red-green-blue (RGB) images using a phenotyping platform based on multi-view stereo. Soybean plants were grown in a controlled indoor environment and imaged repeatedly on Mondays, Wednesdays, and Fridays for three weeks using an RGB camera (LUMIX DMC-G7W, Panasonic, Japan), covering the VC to V2 growth stages. The dataset was originally introduced in Eff-3DPSeg \citep{luo2023eff3dpseg} and was extended in this work with additional reconstructed samples. Manual annotations for stem–leaf semantic segmentation and leaf instance segmentation are available for 145 samples, of which 120 were used for training and 25 for testing, following \citep{luo2023eff3dpseg} and  \citep{tian2025soybeanpcmae}.
\paragraph{HR3D}
HR3D consists of 546 laser-scanned plant point clouds covering three crop species: tomato (312 samples), sorghum (129 samples), and tobacco (105 samples). These samples were captured using a blue-laser scanner (Edge ScanArm HD, FARO, USA). The point clouds were acquired under diverse growth conditions, including ambient light, shade, high heat, high light, and drought, over a period of 20--30 days.
Each sample was down-sampled to 4,096 points and augmented using repeated FPS integration following PlantNet \citep{li2022plantnet}, yielding 3,640 training and 1,820 testing point clouds.
\paragraph{SYAU-Maize}
SYAU-Maize is a maize point cloud dataset consisting of 428 samples from five varieties (Xian Yu 335, LD 145, LD 502, LD586, and LD 1281). The plants were grown in field environments and scanned indoors using a high-precision 3D scanner (FreeScan X3, Tianyuan Inc., China). The dataset contains both complete plant structures and incomplete point clouds caused by acquisition artifacts, which enables the evaluation of segmentation robustness under partial observations.
Following the sample-selection strategy of Deformation3D \citep{yang2024deformation3d}, a training set containing 22 samples was constructed, and the same samples were used for all compared methods.
\paragraph{ShapeNet Part}
ShapeNet Part is a widely-used benchmark for 3D part segmentation, derived from ShapeNetCore 3D CAD models and covering multiple object categories. The dataset contains 16,881 samples from 16 categories, labeled with 50 semantic parts in total. Each sample is labeled with 2 to 5 semantic parts, such as wings, body, and tail for airplanes or seat, back, and legs for chairs.
\subsection{Implementation details}
The pre-training experiments used cross-scale representation learning without manual annotations.
The resulting models were then fine-tuned on downstream plant segmentation tasks under both full and limited supervision, as well as on synthetic part segmentation.
All experiments were implemented in PyTorch.
Pre-training and full-shot fine-tuning were performed on an NVIDIA GeForce RTX 3090 GPU, while few-shot fine-tuning used an NVIDIA GeForce RTX 3060 GPU.
\paragraph{Pre-training}
Pre-training was conducted on the Soybean3D dataset using the same train-test split as \citep{tian2025soybeanpcmae}, yielding 2,890 augmented samples for unsupervised representation learning.
For synthetic part segmentation, pre-training was instead performed on the ShapeNet dataset \citep{shapenet}, which contains 57,448 synthetic 3D shapes from 55 common categories.
Across all pre-training experiments, input point clouds were randomly sampled to 2048 points.
Local neighborhoods were constructed at $S=3$ spatial scales with center counts $\{ N_s\}_{s=1}^3=\{1024,\allowbreak512,\allowbreak128\}$ and $k$-NN sizes $\{ k_s\}_{s=1}^3=\{16,\allowbreak8,\allowbreak8\}$.
Multi-scale masking was applied at a ratio of 0.8 at the coarsest scale, $s=S$.
The encoder followed Point-M2AE \citep{zhang2022pointm2ae}, with encoder embedding dimensions $\{ C_s\}_{s=1}^3=\{96,\allowbreak 192,\allowbreak 384\}$ and five Transformer blocks per stage.
Feature projection at the $s$-th scale used a two-layer MLP $[C_s,\allowbreak 128,\allowbreak D]$ for cross-scale consistency learning.
The lightweight decoder had the same number of stages $S=3$ and used one block per stage with dimension $D=384$.

Pre-training was configured for up to 400 epochs using AdamW \citep{loshchilov2018adamw} with an initial learning rate of $10^{-3}$ and a weight decay of 0.05.
A linear warmup was applied for the first 40 epochs starting from $10^{-6}$, followed by cosine decay with a minimum learning rate of $10^{-6}$.
A batch size of 48 was used for both datasets.
Following Point-M2AE \citep{zhang2022pointm2ae}, early stopping terminated the adopted runs at epoch 300.
\paragraph{Plant Segmentation}
The limited-label protocol was defined at the sample level, with each few-shot sample fully annotated for both semantic and instance segmentation.
This sample-level protocol avoided the annotation effort and sampling sensitivity associated with selecting sparse point labels across plant organs.
For downstream plant segmentation, Soybean3D and SYAU-Maize were preprocessed to 16,384 points using 3DEPS \citep{li2022plantnet}, and non-test samples were expanded tenfold by repeated FPS integration.
The released 4,096-point HR3D samples were used directly.
PlantC2USeg used 4,096 input points, whereas each comparator retained its method-specific input point count.
Before metric computation, all predictions were mapped by nearest-neighbor interpolation to the original-resolution point clouds for Soybean3D and SYAU-Maize and to the released evaluation point clouds for HR3D.
The propagated feature dimension was fixed to $F=384$, and the semantic and offset heads used MLP hidden dimensions $[96,\allowbreak 48]$ with a dropout rate of $0.5$.
For multi-species fine-tuning, a learned category encoding was appended to the global feature context supplied to both prediction heads after $\mathbf{F}$ was formed.

Progressive grouping used five radial steps and initial-radius ratios of $0.7$, $0.6$, and $0.5$ for full-shot, 20-shot, and 10-shot settings, respectively.
Radial progression was disabled for narrow-leaf species by setting the ratio to $0$.
In limited-supervision settings, including the 22-sample SYAU-Maize setting, threshold statistics were accumulated during fine-tuning with an EMA decay of $0.99$, whereas full-shot inference used the non-EMA values saved during training.
Median aggregation was used for the reconnection threshold across all supervision settings.
The DBSCAN and consolidation aggregation quantiles were reduced from $0.5$ and $0.99$ under full supervision to $0$ and $0.9$ under limited supervision, respectively.
For HR3D, threshold values were maintained separately for each species.

PlantC2USeg was optimized using AdamW with a base learning rate of $2\times10^{-4}$ followed by cosine decay.
Full-shot schedules used a weight decay of $0.05$, with warm-up periods of 50 and 25 epochs and maximum durations of 1500 and 800 epochs for Soybean3D and HR3D, respectively.
Limited-supervision schedules reduced the weight decay to $0.01$ and extended training to 2.5 and 5 times the full-shot duration for 20-shot and 10-shot settings, respectively, with longer warm-up periods to mitigate underfitting.
The encoder was unfrozen immediately for Soybean3D few-shot fine-tuning and after 20 epochs for full-shot training and cross-species transfer to HR3D and SYAU-Maize.
The 22-sample SYAU-Maize schedule used a weight decay of $0.01$, a 200-epoch warm-up, and a maximum duration of 6000 epochs.
\paragraph{Object Part Segmentation}
For part segmentation, the same parameters and training protocol as Point-M2AE\linebreak \citep{zhang2022pointm2ae} were adopted.
\subsection{Evaluation metrics}
\paragraph{Semantic segmentation metrics}
Precision (Prec), Recall (Rec), F1-score (F1), and Intersection over Union (IoU) were used to measure the semantic segmentation performance at the point level for each semantic class:
\begin{equation}
\mathrm{Prec}
=
\frac{\mathrm{TP}}{\mathrm{TP} + \mathrm{FP}}
\end{equation}
\begin{equation}
\mathrm{Rec}
=
\frac{\mathrm{TP}}{\mathrm{TP} + \mathrm{FN}}
\end{equation}
\begin{equation}
\mathrm{F1}
=
\frac{2 \cdot \mathrm{Prec} \cdot \mathrm{Rec}}{\mathrm{Prec} + \mathrm{Rec}}
\end{equation}
\begin{equation}
\mathrm{IoU}
=
\frac{\mathrm{TP}}{\mathrm{TP} + \mathrm{FP} + \mathrm{FN}}
\end{equation}
where $\mathrm{TP}$, $\mathrm{FP}$, and $\mathrm{FN}$ denote the numbers of true positive, false positive, and false negative points, respectively.
Precision and recall characterize prediction reliability and ground-truth coverage, respectively.
The F1-score is the harmonic mean of precision and recall, and serves to summarize their trade-off.
IoU quantifies the overlap between the predicted and ground-truth regions for each semantic category.
\paragraph{Instance segmentation metrics}
Instance segmentation performance was evaluated using mean precision (mPrec), mean recall (mRec), mean coverage (mCov), and mean weighted coverage (mWCov).
\begin{equation}
\mathrm{mPrec}
=
\frac{1}{\left| \mathcal{Y}_{\mathrm{inst}} \right|}
\sum_{y \in \mathcal{Y}_{\mathrm{inst}}}
\frac{
\left| \mathrm{TP}^{(y)}_{\mathrm{ins}} \right|
}{
\left| \hat{\mathcal{I}}^{(y)} \right|
}
\qquad
\end{equation}
\begin{equation}
\mathrm{mRec}
=
\frac{1}{\left| \mathcal{Y}_{\mathrm{inst}} \right|}
\sum_{y \in \mathcal{Y}_{\mathrm{inst}}}
\frac{
\left| \mathrm{TP}^{(y)}_{\mathrm{ins}} \right|
}{
\left| \mathcal{I}^{(y)} \right|
}
\end{equation}
where $\mathcal{Y}_{\mathrm{inst}} \subset \mathcal{Y}$ denotes the set of instance-aware semantic categories.
For each category $y \in \mathcal{Y}_{\mathrm{inst}}$, $\mathcal{I}^{(y)}$ and $\hat{\mathcal{I}}^{(y)}$ represent the sets of ground-truth and predicted instances, respectively.
$\mathrm{TP}^{(y)}_{\mathrm{ins}}$ is defined as the number of predicted instances in category $y$ whose maximum IoU with any ground-truth instance in the same category exceeds $0.5$.
Mean coverage (mCov) was computed per instance-aware semantic category and then averaged across categories:
\begin{equation}
\mathrm{mCov}
=
\frac{1}{|\mathcal{I}^{(y)}|}
\sum_{\ell \in \mathcal{I}^{(y)}}
\max_{j}
\mathrm{IoU}
\left(
\mathcal{A}_\ell,\, \hat{\mathcal{A}}_{j}
\right)
\end{equation}
where $\mathcal{A}_\ell$ denotes the set of points belonging to the $\ell$-th ground-truth instance.
For each instance $\ell$, the IoU is defined as the maximum overlap between the ground-truth $\mathcal{A}_\ell$ and the predicted instance point sets $\hat{\mathcal{A}}_j$ in the same semantic category.

Mean weighted coverage (mWCov) extends mCov by introducing instance-size weighting:
\begin{equation}
\mathrm{mWCov}
=
\sum_{\ell \in \mathcal{I}^{(y)}}
\omega_\ell
\max_{j}
\mathrm{IoU}
\left(
\mathcal{A}_\ell,\, \hat{\mathcal{A}}_{j}
\right)
\qquad
\end{equation}
\begin{equation}
\omega_\ell
=
\frac{
|\mathcal{A}_\ell|
}{
\sum_{\ell' \in \mathcal{I}^{(y)}}
|\mathcal{A}_{\ell'}|
}
\end{equation}
where the weight $\omega_\ell$ is proportional to$|\mathcal{A}_\ell|$, such that instances with more points contribute more to the overall mWCov score.
mWCov emphasizes spatial coverage on dominant plant structures and is less sensitive to small fragmented instances.
\paragraph{Part segmentation metrics}
Following common practice on the ShapeNet Part dataset, two variants of mean IoU were adopted, including category-averaged mIoU and instance-averaged mIoU.
The category-averaged mIoU averages IoU scores over all part categories, while the instance-averaged mIoU averages IoU scores over all samples.

For all metrics, higher values indicate better segmentation performance.

\section{Results}
\subsection{Plant stem--leaf segmentation}
\subsubsection{Soybean3D}
\begin{figure*}[pos=htbp]
\centering
\includegraphics[width=\textwidth]{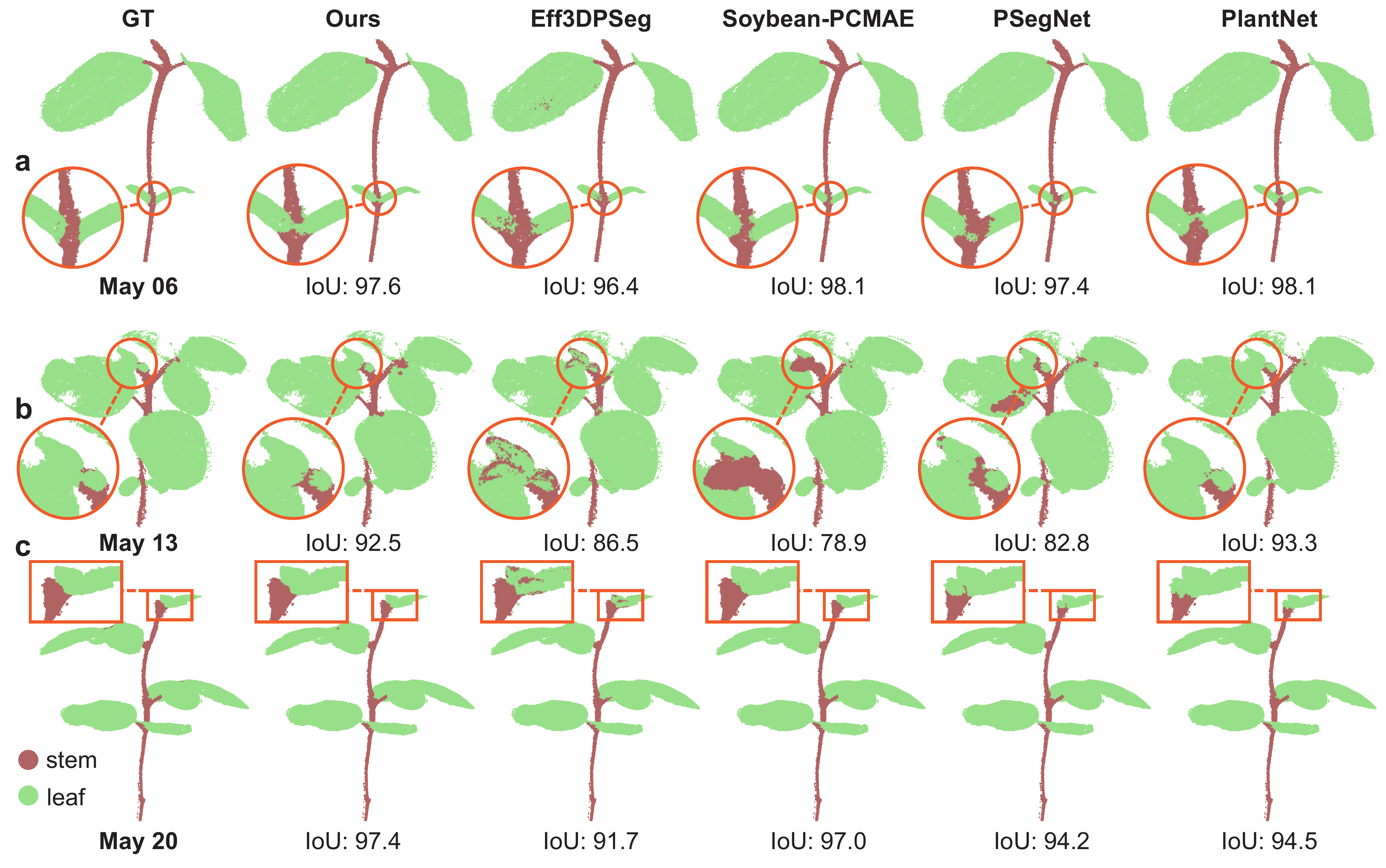}
\caption{Rows show one Soybean3D plant imaged on May 06, May 13, and May 20 from top to bottom.
Columns from left to right are GT (ground-truth reference) followed by the full-shot stem--leaf semantic predictions from Ours (PlantC2USeg), Eff3DPSeg, Soybean-PCMAE, PSegNet, and PlantNet, with displayed values denoting sample-level IoU (\%).
Regions~a, b, and c mark a narrow stem--leaf junction, a small apical leaf, and stem continuation beside a leaf, respectively.}\label{fig6}
\end{figure*}

Full-shot stem--leaf semantic predictions for one soybean plant imaged on May 06, May 13, and May 20 differed in class allocation at narrow stem--leaf junctions (Figure~\ref{fig6}).
In region~b on May 13, PlantC2USeg and PlantNet retained more of the ground-truth small-leaf area, whereas Eff3DPSeg, Soybean-PCMAE, and PSegNet showed greater leaf-to-stem leakage.
On the May 13 sample, PlantC2USeg and PlantNet reached 92.5\% and 93.3\% IoU, respectively, whereas Eff3DPSeg, Soybean-PCMAE, and PSegNet remained below 87\%.
In region~c on May 20, PlantNet misclassified part of the ground-truth stem as leaf, whereas PlantC2USeg preserved a continuous visible stem path.
For PlantNet, regions~b and~c contrast small-leaf retention with stem-to-leaf leakage.

Dataset-level evaluation on Soybean3D covered stem--leaf semantic and leaf instance segmentation under full-shot, 20-shot, and 10-shot supervision (Table~\ref{table1}), with the instance results analyzed later in Section~\ref{sec:soybean3d-leaf-instance}.
Under full supervision, PlantC2USeg combined cross-scale pre-training with joint fine-tuning to achieve the highest semantic F1 and IoU, reaching 95.69\% and 91.91\%, respectively.
Among the pre-trained alternatives that adapt the two tasks separately, Soybean-PCMAE \citep{tian2025soybeanpcmae} uses staged fine-tuning and reached 90.97\% IoU, whereas Eff-3DPSeg \citep{luo2023eff3dpseg} trains separate task networks and reached 88.41\%.
PSegNet \citep{li2022psegnet}, a fully supervised joint baseline with dual-scale features, reached 89.29\% IoU.
With the downstream architecture and joint fine-tuning setup held fixed, pre-training increased IoU by 1.72 percentage points.
The increase was driven mainly by recall, which rose from 92.98\% to 95.01\%, while precision remained similar at 96.64\% for the scratch model and 96.40\% for PlantC2USeg.
\begin{table*}
\centering
\caption{Quantitative results of stem--leaf and leaf instance segmentation on the Soybean3D dataset (\%). A checkmark denotes joint optimization of stem--leaf semantic and leaf instance segmentation within a single downstream model. H. Transformer denotes the Hierarchical Transformer trained from scratch. The best results are in boldface, and the 2nd best results are underlined.}\label{table1}
\begin{tabular}{lccccccccc}
\hline
Method & Joint
& \multicolumn{4}{c}{Semantic Segmentation}
& \multicolumn{4}{c}{Instance Segmentation} \\
\cmidrule(lr){3-6} \cmidrule(lr){7-10}
 & & Prec & Rec & F1 & IoU & mCov & mWCov & mPrec & mRec \\
\hline
\multicolumn{10}{l}{\textit{Full-shot}}\\
PlantNet & \checkmark & 95.13 & 91.30 & 93.09 & 87.50 & 77.53 & 87.40 & 88.29 & 76.86\\
PSegNet & \checkmark & 95.78 & 92.71 & 94.17 & 89.29 & 81.30 & 87.89 & 89.92 & 83.92\\
Eff-3DPSeg & & 94.80 & 92.58 & 93.65 & 88.41 & 81.51 & 88.90 & 79.78 & 86.67\\
Soybean-PCMAE & & 95.95 & \underline{94.40} & \underline{95.15} & \underline{90.97} & \underline{85.85} & 91.90 & 89.96 & \underline{91.37}\\
H. Transformer & \checkmark & \textbf{96.64} & 92.98 & 94.70 & 90.19 & 85.54 & \underline{92.11} & \underline{94.56} & 88.63\\
PlantC2USeg & \checkmark & \underline{96.40} & \textbf{95.01} & \textbf{95.69} & \textbf{91.91} & \textbf{89.94} & \textbf{94.62} & \textbf{96.39} & \textbf{94.12}\\
\hline
\multicolumn{10}{l}{\textit{20-shot}}\\
Soybean-PCMAE & & 93.50 & \textbf{95.06} & \underline{94.26} & \underline{89.42} & 77.03 & 84.56 & 90.87 & \underline{81.96}\\
Deformation3D & & 88.23 & 89.27 & 88.74 & 80.70 & \underline{79.41} & \underline{87.43} & \underline{90.91} & 78.43\\
H. Transformer & \checkmark & \underline{94.20} & 91.68 & 92.89 & 87.15 & 74.47 & 82.26 & 75.09 & 79.22\\
PlantC2USeg & \checkmark & \textbf{94.31} & \underline{94.63} & \textbf{94.47} & \textbf{89.78} & \textbf{83.94} & \textbf{90.27} & \textbf{92.71} & \textbf{89.80}\\
\hline
\multicolumn{10}{l}{\textit{10-shot}}\\
Soybean-PCMAE & & \underline{89.88} & \textbf{93.14} & \textbf{91.41} & \textbf{84.75} & \underline{72.28} & 75.92 & \underline{79.03} & \textbf{79.67}\\
H. Transformer & \checkmark & 88.98 & \underline{90.21} & 89.58 & 81.95 & 72.13 & \underline{79.49} & 75.00 & 75.29\\
PlantC2USeg & \checkmark & \textbf{90.72} & 90.06 & \underline{90.39} & \underline{83.19} & \textbf{75.76} & \textbf{83.23} & \textbf{83.40} & \underline{78.82}\\
\hline
\end{tabular}
\end{table*}

\begin{figure*}[pos=htbp]
\centering
\includegraphics[width=\textwidth]{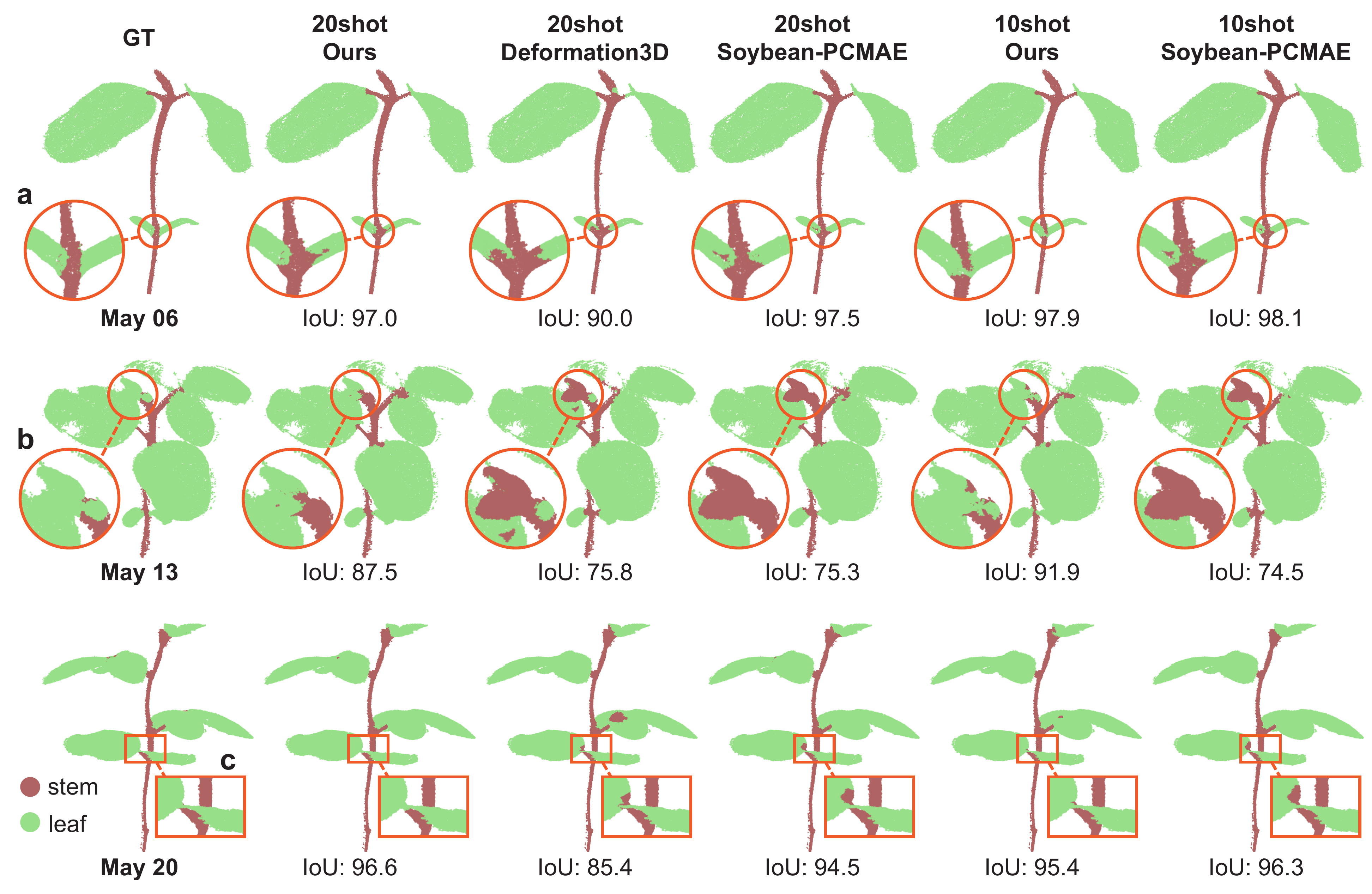}
\caption{Few-shot stem--leaf semantic segmentation of the May 06, May 13, and May 20 Soybean3D samples reused from Figure~\ref{fig6}.
Columns from left to right show ground truth (GT), 20-shot PlantC2USeg, 20-shot Deformation3D, 20-shot Soybean-PCMAE, 10-shot PlantC2USeg, and 10-shot Soybean-PCMAE.
The May 06, May 13, and May 20 rows contain regions~a, b, and c, respectively, marking boundary allocation at a narrow junction, retention of small leaves at the compact apex, and leaf-to-stem leakage beside a junction.
Red denotes stem and green denotes leaf.
Displayed values are sample-level mean IoU over stem and leaf (\%).}\label{fig7}
\end{figure*}

Figure~\ref{fig7} extends the same three Soybean3D examples from Figure~\ref{fig6} to 20-shot and 10-shot stem--leaf semantic segmentation.
In region~b of the May 13 sample, PlantC2USeg retained more of the ground-truth small-leaf area as leaf, with less leaf-to-stem leakage than Soybean-PCMAE at both budgets and Deformation3D at 20 shots.
For the complete May 13 sample, PlantC2USeg exceeded every alternative shown at the same annotation budget by at least 11.7 percentage points in sample-level IoU.
In region~c on May 20, PlantC2USeg preserved more of the ground-truth leaf tissue adjoining the stem, whereas Soybean-PCMAE at both budgets and 20-shot Deformation3D extended the stem label into this area.

At the dataset level, PlantC2USeg retained the highest semantic F1 and IoU under 20-shot supervision on Soybean3D (Table~\ref{table1}), reaching 94.47\% and 89.78\%, respectively.
The 20-shot IoU exceeded the full-shot results of PlantNet \citep{li2022plantnet}, PSegNet \citep{li2022psegnet}, and Eff-3DPSeg \citep{luo2023eff3dpseg} by 0.49--2.28 percentage points, and F1 was also higher across this group.
At the same annotation count, PlantC2USeg exceeded Deformation3D \citep{yang2024deformation3d} by 9.08 percentage points in IoU, supporting representation transfer over deformation-based adaptation in this comparison.
With the downstream setup fixed, pre-training increased IoU over the Hierarchical Transformer trained from scratch by 2.63 percentage points at 20 shots and 1.24 percentage points at 10 shots, extending the full-shot benefit across supervision levels.
At 10 shots, Soybean-PCMAE \citep{tian2025soybeanpcmae} reached 84.75\% IoU compared with 83.19\% for PlantC2USeg and was also higher in F1 and recall.
PlantC2USeg retained the highest precision at 90.72\% among the three 10-shot methods.

\subsubsection{HR3D}
\begin{figure*}[pos=htbp]
\centering
\includegraphics[width=\textwidth]{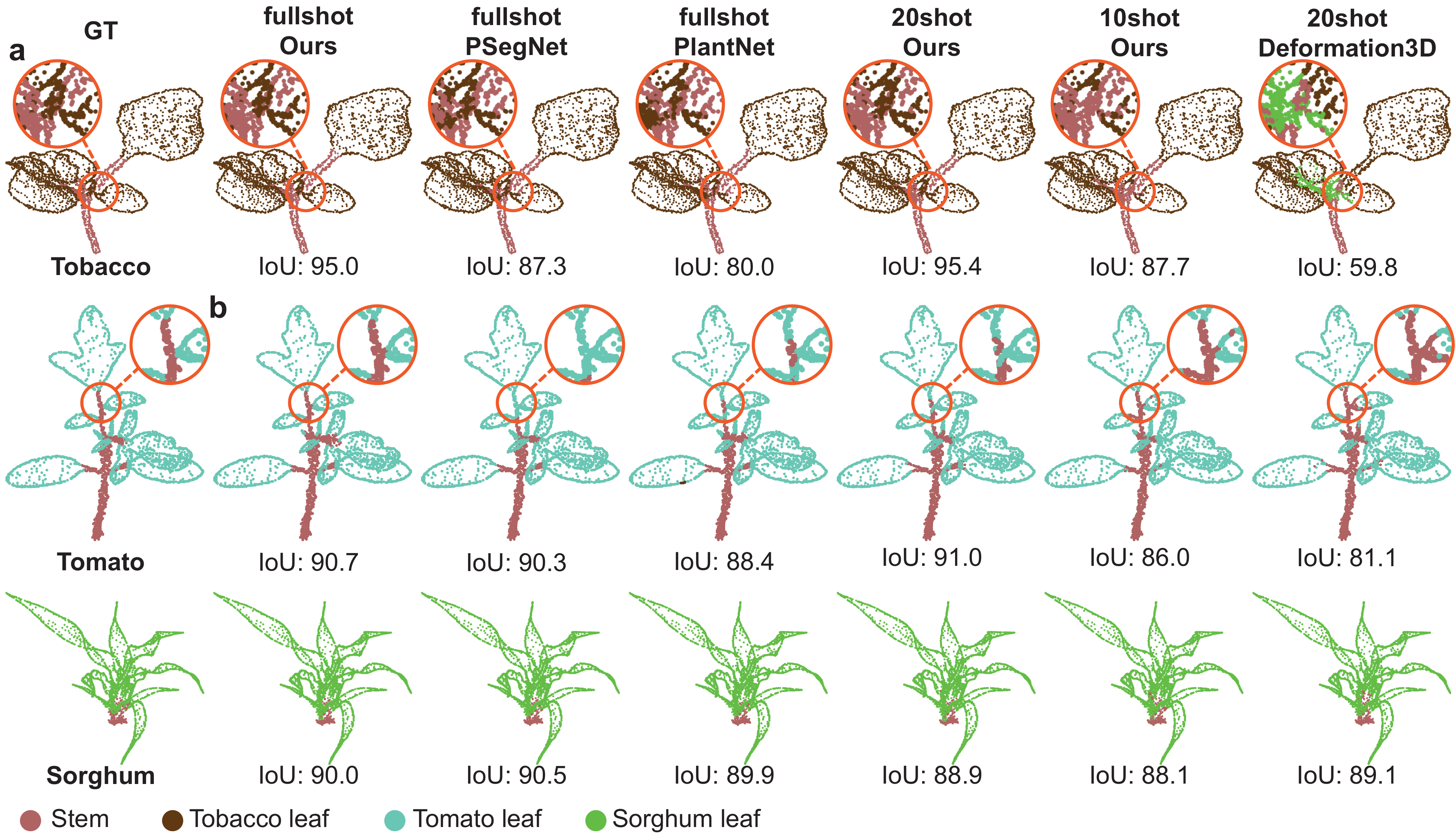}
\caption{Selected HR3D stem--leaf semantic segmentation results, with tobacco, tomato, and sorghum shown from top to bottom.
Columns from left to right show ground truth (GT), the full-shot predictions of PlantC2USeg, PSegNet, and PlantNet, the 20-shot and 10-shot predictions of PlantC2USeg, and the 20-shot Deformation3D prediction.
The legend identifies stem and the species-specific leaf classes.
Region~a marks an emerging tobacco leaf, while region~b marks a narrow tomato stem.
At the unmarked sorghum base, all methods showed local disagreement with ground truth without a consistent error direction.
Displayed values are sample-level mean IoU over stem and the corresponding species leaf class (\%).}\label{fig8}
\end{figure*}

Figure~\ref{fig8} compares full-shot and few-shot stem--leaf predictions on selected tobacco, tomato, and sorghum samples from HR3D.
In region~a of the tobacco example, full-shot and 20-shot PlantC2USeg recovered the complete emerging leaf.
Full-shot PSegNet and PlantNet recovered only the part of the emerging leaf adjoining the left leaf, while 10-shot PlantC2USeg recovered less of the same structure than either baseline.
The 20-shot Deformation3D model instead assigned the emerging leaf to the sorghum leaf class.
Region~b of the tomato example showed bidirectional leakage around a narrow stem.
Full-shot PSegNet assigned the entire narrow stem to the tomato leaf class, while full-shot PlantNet and 20-shot PlantC2USeg showed partial stem-to-leaf leakage.
Both 10-shot PlantC2USeg and 20-shot Deformation3D extended the stem class into surrounding leaves.
Full-shot PlantC2USeg preserved the complete stem.

HR3D extends the dataset-level semantic evaluation from soybean to tobacco, tomato, and sorghum under full-shot, 20-shot, and 10-shot supervision (Table~\ref{table2}).
Leaf instance results from the same evaluation are reported in the table and examined in Section~\ref{sec:hr3d-leaf-instance}.
Under full supervision, PlantC2USeg exceeded PSegNet \citep{li2022psegnet}, the strongest plant-specific joint comparator in Table~\ref{table2}, by 1.80 percentage points in aggregate IoU and achieved higher F1 and IoU for all three species.
For tomato, PlantC2USeg paired 0.75 percentage points higher recall with 0.22 percentage points lower precision than PSegNet, whereas\linebreak sorghum showed the reverse direction of higher precision and lower recall.
Despite these opposing precision--recall patterns, PlantC2USeg achieved higher F1 and IoU for both species.

At 20 shots, PlantC2USeg transfers a pre-trained representation, whereas Deformation3D \citep{yang2024deformation3d} uses deformation-based adaptation from the same 20 original annotated HR3D samples.
PlantC2USeg reached 81.50\% aggregate IoU, 15.38 percentage points above Deformation3D.
Across species, the difference ranged from 7.95 percentage points for tomato to 19.80 for tobacco and was also large for sorghum.
The annotation-efficiency advantage of transferring a pre-trained representation over deformation-based adaptation thus persisted across the evaluated HR3D species.

Against the full-shot general-purpose joint methods SGPN \citep{wang2018sgpn} and ASIS \citep{wang2019asis}, transfer of a pre-trained representation preserved competitive semantic performance with substantially fewer labels.
The 20-shot and 10-shot PlantC2USeg models exceeded SGPN by 13.55 and 10.46 percentage points in aggregate IoU, respectively.
The 20-shot model was also within 1.50 percentage points of ASIS in IoU, with a similarly close F1, comparable recall, and lower precision.
\begin{table*}
\centering
\caption{Quantitative results of stem--leaf and leaf instance segmentation on HR3D (\%). Sup. denotes supervision. Full uses the complete training set, and numerical entries give the number of annotated training samples. A checkmark denotes joint optimization of stem--leaf semantic and leaf instance segmentation within a single downstream model.}\label{table2}
\begin{tabular}{lcccccccccc}
\hline
Method & Sup. & Joint
& \multicolumn{4}{c}{Semantic Segmentation}
& \multicolumn{4}{c}{Instance Segmentation} \\
\cmidrule(lr){4-7} \cmidrule(lr){8-11}
 & & & Prec & Rec & F1 & IoU & mCov & mWCov & mPrec & mRec \\
\hline
\multicolumn{11}{l}{\textit{Mean}}\\
SGPN & Full & \checkmark & 82.08 & 76.48 & 76.61 & 67.95 & 41.68 & 48.78 & 40.66 & 30.14\\
ASIS & Full & \checkmark & 91.31 & 88.57 & 89.18 & 83.00 & 70.39 & 77.45 & 78.66 & 62.96\\
PlantNet & Full & \checkmark & 91.86 & 91.45 & 90.98 & 85.10 & 76.17 & 81.68 & 83.56 & 72.61\\
PSegNet & Full & \checkmark & \underline{93.37} & \underline{92.34} & \underline{92.27} & \underline{87.27} & 79.39 & 83.67 & \underline{86.58} & 73.70\\
PlantC2USeg & Full & \checkmark & \textbf{94.13} & \textbf{93.19} & \textbf{93.37} & \textbf{89.07} & \textbf{87.78} & \textbf{91.58} & \textbf{89.00} & \textbf{86.88}\\
Deformation3D & 20 & & 80.79 & 73.30 & 73.83 & 66.12 & 66.67 & 73.91 & 78.92 & 67.12\\
PlantC2USeg & 20 & \checkmark & 88.24 & 88.77 & 87.43 & 81.50 & \underline{81.20} & \underline{86.21} & 80.28 & \underline{83.86}\\
PlantC2USeg & 10 & \checkmark & 85.45 & 86.97 & 85.14 & 78.41 & 72.73 & 79.42 & 74.06 & 70.16\\

\cline{1-11}
\multicolumn{11}{l}{\textit{Tobacco}}\\
SGPN & Full & \checkmark & 79.51 & 73.29 & 74.05 & 62.20 & 28.65 & 37.66 & 25.41 & 22.70\\
ASIS & Full & \checkmark & 89.88 & 88.42 & 88.66 & 82.31 & 64.08 & 74.65 & 77.59 & 60.04\\
PlantNet & Full & \checkmark & 90.76 & 89.01 & 89.08 & 82.35 & 70.84 & 79.85 & 84.56 & 68.37\\
PSegNet & Full & \checkmark & \underline{93.45} & \underline{90.78} & \underline{91.39} & \underline{86.23} & 76.82 & 84.00 & \underline{89.11} & 72.80\\
PlantC2USeg & Full & \checkmark & \textbf{94.34} & \textbf{94.54} & \textbf{94.18} & \textbf{90.14} & \textbf{86.12} & \textbf{92.19} & \textbf{89.33} & \underline{83.44}\\
Deformation3D & 20 & & 79.24 & 72.90 & 73.60 & 64.73 & 65.28 & 75.77 & 83.70 & 65.92\\
PlantC2USeg & 20 & \checkmark & 91.30 & 90.18 & 89.98 & 84.53 & \underline{83.98} & \underline{89.88} & 83.05 & \textbf{85.99}\\
PlantC2USeg & 10 & \checkmark & 89.84 & 90.16 & 89.64 & 82.98 & 76.44 & 85.29 & 75.45 & 74.01\\

\cline{1-11}
\multicolumn{11}{l}{\textit{Tomato}}\\
SGPN & Full & \checkmark & 86.54 & 83.36 & 83.55 & 77.10 & 49.57 & 58.77 & 76.05 & 30.94\\
ASIS & Full & \checkmark & 94.25 & 93.16 & 93.45 & 88.42 & 76.27 & 81.77 & 84.48 & 64.95\\
PlantNet & Full & \checkmark & 94.91 & 94.04 & 94.32 & 89.87 & 80.88 & 84.97 & 85.75 & 73.68\\
PSegNet & Full & \checkmark & \textbf{95.95} & \underline{94.90} & \underline{95.30} & \underline{91.45} & \underline{81.68} & \underline{85.10} & \underline{88.30} & 72.40\\
PlantC2USeg & Full & \checkmark & \underline{95.73} & \textbf{95.65} & \textbf{95.58} & \textbf{91.95} & \textbf{87.65} & \textbf{90.69} & \textbf{91.34} & \textbf{86.15}\\
Deformation3D & 20 & & 90.63 & 86.24 & 86.63 & 79.42 & 70.86 & 78.17 & 86.87 & 68.03\\
PlantC2USeg & 20 & \checkmark & 93.09 & 92.53 & 92.58 & 87.37 & 80.58 & 84.69 & 78.47 & \underline{79.73}\\
PlantC2USeg & 10 & \checkmark & 92.22 & 91.55 & 91.60 & 85.53 & 72.95 & 78.65 & 69.59 & 61.88\\

\cline{1-11}
\multicolumn{11}{l}{\textit{Sorghum}}\\
SGPN & Full & \checkmark & 80.17 & 72.79 & 72.23 & 64.55 & 46.83 & 49.93 & 20.51 & 36.78\\
ASIS & Full & \checkmark & 89.80 & 84.13 & 85.44 & 78.27 & 70.84 & 75.95 & 73.90 & 63.91\\
PlantNet & Full & \checkmark & 89.91 & \underline{91.30} & 89.54 & 83.07 & 76.79 & 80.21 & 80.37 & 75.79\\
PSegNet & Full & \checkmark & \underline{90.70} & \textbf{91.34} & \underline{90.10} & \underline{84.13} & \underline{79.67} & 81.90 & \underline{82.33} & 75.89\\
PlantC2USeg & Full & \checkmark & \textbf{92.32} & 89.38 & \textbf{90.34} & \textbf{85.12} & \textbf{89.58} & \textbf{91.86} & \textbf{86.34} & \textbf{91.04}\\
Deformation3D & 20 & & 72.50 & 60.76 & 61.27 & 54.19 & 63.87 & 67.80 & 66.18 & 67.43\\
PlantC2USeg & 20 & \checkmark & 80.32 & 83.59 & 79.74 & 72.60 & 79.05 & \underline{84.07} & 79.32 & \underline{85.84}\\
PlantC2USeg & 10 & \checkmark & 74.31 & 79.21 & 74.19 & 66.72 & 68.81 & 74.32 & 77.12 & 74.60\\

\hline
\end{tabular}
\end{table*}

\subsubsection{SYAU-Maize}
\begin{figure*}[pos=htbp]
\centering
\includegraphics{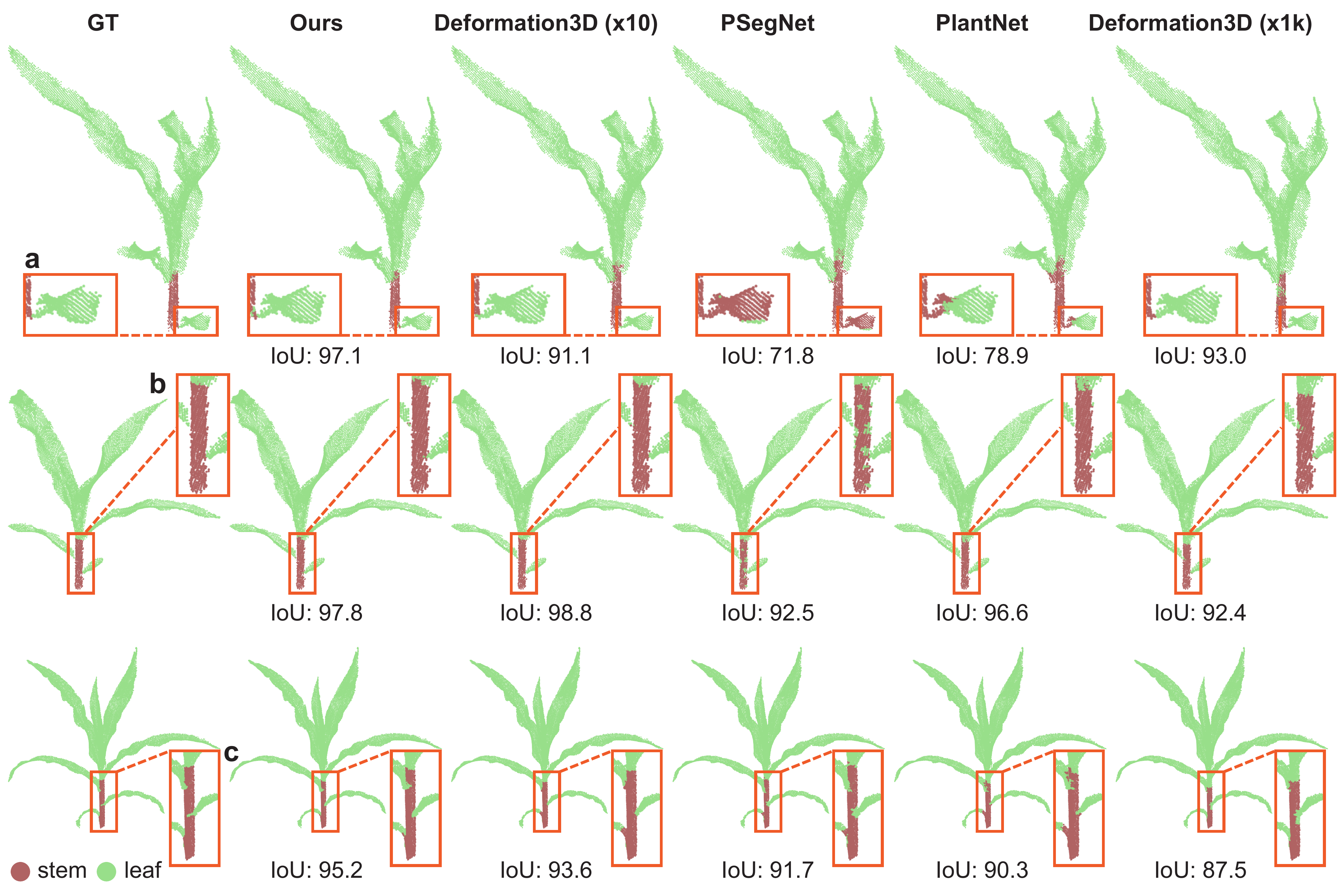}
\caption{Stem--leaf semantic segmentation on three SYAU-Maize samples under a 22-sample annotation budget, with displayed values denoting sample-level IoU (\%).
Columns from left to right show ground truth (GT), PlantC2USeg, Deformation3D (x10), PSegNet, PlantNet, and Deformation3D (x1k).
The x10 and x1k settings expand the same 22-sample training set to 220 and 22,\allowbreak000 deformed point clouds, respectively.
Region~a marks a cotyledon near abscission, and regions~b and~c mark collar boundaries adjoining the leaf sheath.}\label{fig9}
\end{figure*}

Figure~\ref{fig9} compares semantic predictions under the 22-sample Deformation3D protocol \citep{yang2024deformation3d}, focusing on leaf-to-stem errors at a cotyledon near abscission and in leaf-sheath tissue at the collar.
The cotyledon in region~a was assigned almost entirely to stem by PSegNet and only at the portion adjoining the stem by PlantNet, whereas PlantC2USeg and both Deformation3D variants retained the organ as leaf.
Regions~b and~c show the adjoining leaf sheath closely appressed to the culm at the collar.
PlantC2USeg followed the ground-truth collar edge in region~b more closely than PSegNet, which showed the clearest stem expansion into this locally stem-like tissue.
The PlantC2USeg boundary in region~c remained closer to ground truth than those of all four alternatives, which extended stem assignments into leaf tissue at the collar.

SYAU-Maize provides a shared sparse-label test of stem--leaf semantic segmentation in which every method begins with the same 22 annotated source samples (Table~\ref{table3}).
The table also reports leaf instance segmentation, which is examined in Section~\ref{sec:syau-maize-leaf-instance}.
Under this shared protocol, PlantC2USeg adapted a pre-trained representation through joint fine-tuning, whereas the fully supervised joint methods PlantNet \citep{li2022plantnet} and PSegNet \citep{li2022psegnet} were trained directly from the available labels.
PlantC2USeg reached 92.75\% semantic IoU, 8.94 and 9.45 percentage points above PlantNet and PSegNet, respectively, and also led both methods in precision, recall, and F1.
Within the evaluated sparse-label protocol, the lead across all four semantic metrics indicates that adapting a pre-trained representation was more effective than training these two joint baselines directly from the available labels.

The Deformation3D \citep{yang2024deformation3d} variants showed a different precision--recall balance, with x10 slightly higher than PlantC2USeg in recall and x1k slightly higher in precision.
PlantC2USeg remained higher in F1 and IoU than both variants, reaching 92.75\% IoU.
From x10 to x1k, the generated data volume increased 100-fold as precision rose and recall fell.
F1 and IoU also declined, with IoU decreasing from 91.19\% to 90.59\%.
The additional burden of generating and processing data therefore did not translate into an IoU gain.
\begin{table*}
\centering
\caption{Dataset-level stem--leaf semantic and leaf instance segmentation results on the 406-sample SYAU-Maize test set (\%).
All methods use the same 22 annotated source samples.
For Deformation3D, x10 and x1k denote tenfold and 1000-fold deformation expansion of this training set.
A checkmark denotes joint optimization of both tasks within a single downstream model.}\label{table3}
\begin{tabular}{lccccccccc}
\hline
Method & Joint
& \multicolumn{4}{c}{Semantic Segmentation}
& \multicolumn{4}{c}{Instance Segmentation} \\
\cmidrule(lr){3-6} \cmidrule(lr){7-10}
& & Prec & Rec & F1 & IoU & mCov & mWCov & mPrec & mRec \\
\hline
PlantNet & \checkmark & 91.21 & 90.99 & 90.10 & 83.81 & 67.16 & 72.73 & 84.17 & 71.70\\
PSegNet & \checkmark & 90.70 & 91.00 & 89.78 & 83.30 & 66.09 & 71.78 & 83.03 & 70.40\\
Deformation3D (x10) & & 94.61 & \textbf{96.20} & \underline{95.07} & \underline{91.19} & 84.93 & 91.52 & \underline{91.33} & 85.34\\
Deformation3D (x1k) & & \textbf{97.27} & 92.90 & 94.72 & 90.59 & \underline{88.23} & \textbf{94.90} & 87.70 & \underline{88.25}\\
PlantC2USeg & \checkmark & \underline{97.03} & \underline{95.40} & \textbf{96.04} & \textbf{92.75} & \textbf{89.11} & \underline{92.23} & \textbf{91.46} & \textbf{93.51}\\
\hline
\end{tabular}
\end{table*}

\subsection{Plant leaf instance segmentation}
\subsubsection{Soybean3D}\label{sec:soybean3d-leaf-instance}
\begin{figure*}[pos=htbp]
\centering
\includegraphics[width=\textwidth]{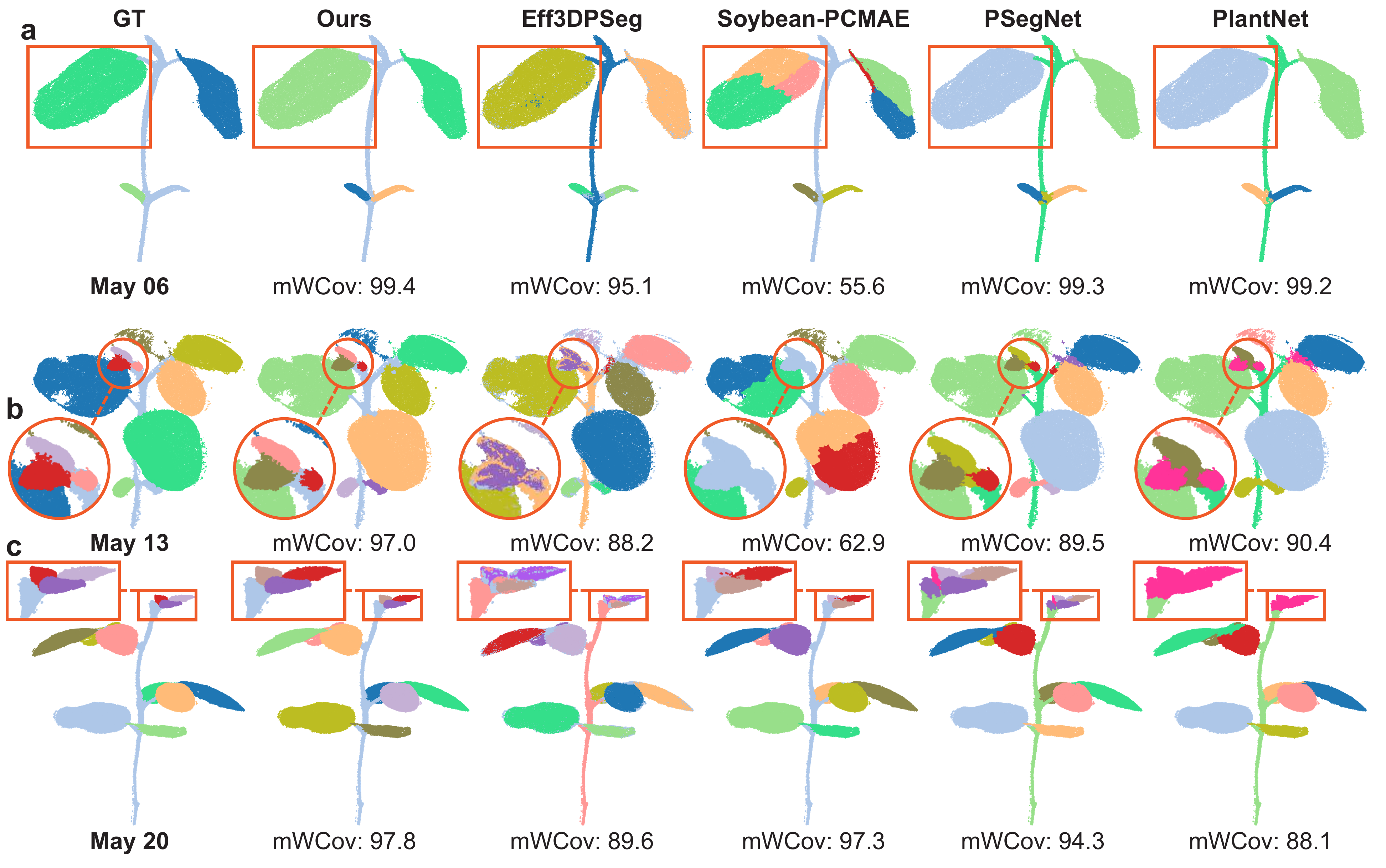}
\caption{Full-shot leaf instance segmentation on three Soybean3D examples.
Regions~a--c mark a broad leaf, three leaves at a compact apex, and three slender apical leaves, respectively.
Values denote sample-level mWCov (\%).}\label{fig10}
\end{figure*}

Figure~\ref{fig10} compares full-shot leaf instance predictions on the same three Soybean3D examples used for the semantic comparison in Figure~\ref{fig6}.
Soybean-PCMAE fragmented the broad leaf in region~a, whereas PlantC2USeg preserved it as one instance.
On the complete May 06 sample, PlantC2USeg reached 99.4\% mWCov, compared with 55.6\% for Soybean-PCMAE.
At the compact apex in region~b, PlantC2USeg and PSegNet separated all three closely grouped leaves, whereas PlantNet merged the two lateral leaves.
Eff3DPSeg did not separate all three leaves, and Soybean-PCMAE failed to recover the corresponding leaf instances.
PlantC2USeg and Soybean-PCMAE separated all three slender apical leaves in region~c.
Eff3DPSeg merged the two rear leaves, PSegNet over-segmented the group into four instances, and PlantNet merged all three.

At the dataset level, full-shot PlantC2USeg led all four Soybean3D instance metrics (Table~\ref{table1}).
Against the plant-specific joint comparator PSegNet \citep{li2022psegnet},\linebreak PlantC2USeg reached 94.62\% mWCov compared with\linebreak 87.89\%.
The mRec margin reached 10.20 percentage points, the largest of the four metric differences, showing that the strongest gain concerned the recovery of ground-truth leaves.
The lead across the coverage and matched-instance measures also extended to Soybean-PCMAE \citep{tian2025soybeanpcmae}.
With the downstream model fixed, pre-training improved coverage and matched-instance performance over the Hierarchical Transformer trained from scratch, including a 5.49 percentage point gain in mRec.

\begin{figure*}[pos=htbp]
\centering
\includegraphics[width=\textwidth]{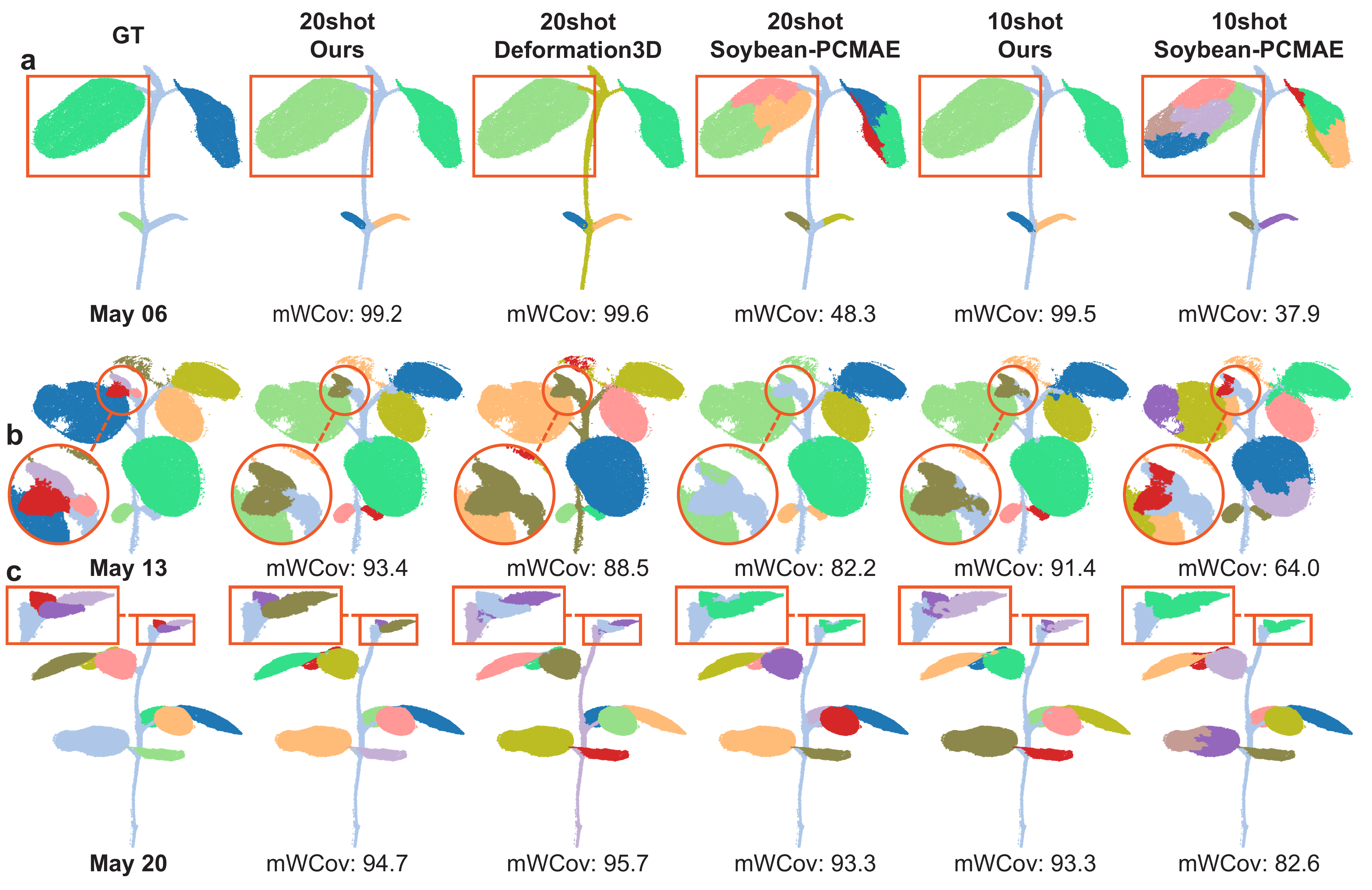}
\caption{Few-shot leaf instance segmentation on the same Soybean3D examples and regions as Figure~\ref{fig10}.
After ground truth (GT), prediction columns show 20-shot PlantC2USeg, Deformation3D, and Soybean-PCMAE, followed by 10-shot PlantC2USeg and Soybean-PCMAE.
Regions~a--c mark the broad leaf, compact apex, and three slender apical leaves used in the full-shot comparison, respectively.
Values denote sample-level mWCov (\%).}\label{fig11}
\end{figure*}

Figure~\ref{fig11} extends the same Soybean3D instance comparison to 20-shot and 10-shot supervision.
Reduced supervision exposed persistent broad-leaf fragmentation for Soybean-PCMAE across all three examples, most clearly in region~a.
PlantC2USeg retained the broad leaf as one instance at both budgets, as did 20-shot Deformation3D, whereas Soybean-PCMAE fragmented it at both budgets and introduced more internal partitions at 10 shots.
At the compact apex in region~b, all displayed predictions remained under-segmented, although PlantC2USeg separated most leaves at both budgets.
Soybean-PCMAE recovered only a small distinct leaf partition, while 20-shot Deformation3D recovered no distinct apical leaf instance.
In region~c, PlantC2USeg separated all three slender leaves at both budgets.
At 20 shots, Deformation3D merged two leaves, while Soybean-PCMAE assigned all three to one instance at both budgets.
Despite this local merge, Deformation3D still had slightly higher mWCov for the complete sample than PlantC2USeg under 20-shot supervision.

At the dataset level, pre-training improved the coverage and matched-instance metrics over the Hierarchical Transformer trained from scratch at both 20 and 10 shots (Table~\ref{table1}).
The mWCov gains ranged from 3.74 to 8.01 percentage points, while mPrec also increased at both supervision levels.
At 20 shots, PlantC2USeg led every method trained with the same annotation count and also surpassed full-shot PlantNet \citep{li2022plantnet}, PSegNet \citep{li2022psegnet}, and Eff-3DPSeg \citep{luo2023eff3dpseg} across the coverage and matched-instance measures.
Its mRec exceeded every named full-shot alternative by at least 3.13 percentage points.
Against Deformation3D \citep{yang2024deformation3d} at the same annotation count, the mRec advantage reached 11.37 percentage points, while mPrec was similar and both coverage measures were higher.

At 10 shots, PlantC2USeg exceeded Soybean-PCMAE \citep{tian2025soybeanpcmae} in both coverage measures and mPrec, with a 7.31 percentage point advantage in mWCov.
Soybean-PCMAE retained only a 0.85 percentage point advantage in mRec.
For semantic segmentation, Soybean-PCMAE had 1.56 percentage points higher IoU and was also higher in F1 and recall, while PlantC2USeg retained the highest precision.
Taken across both tasks, Soybean-PCMAE, which adapts the objectives in stages, retained stronger semantic F1, IoU, and recall together with a narrow mRec advantage, whereas the jointly fine-tuned PlantC2USeg model retained stronger instance coverage and mPrec together with the highest semantic precision.

\subsubsection{HR3D}\label{sec:hr3d-leaf-instance}
\begin{figure*}[pos=htbp]
\centering
\includegraphics[width=\textwidth]{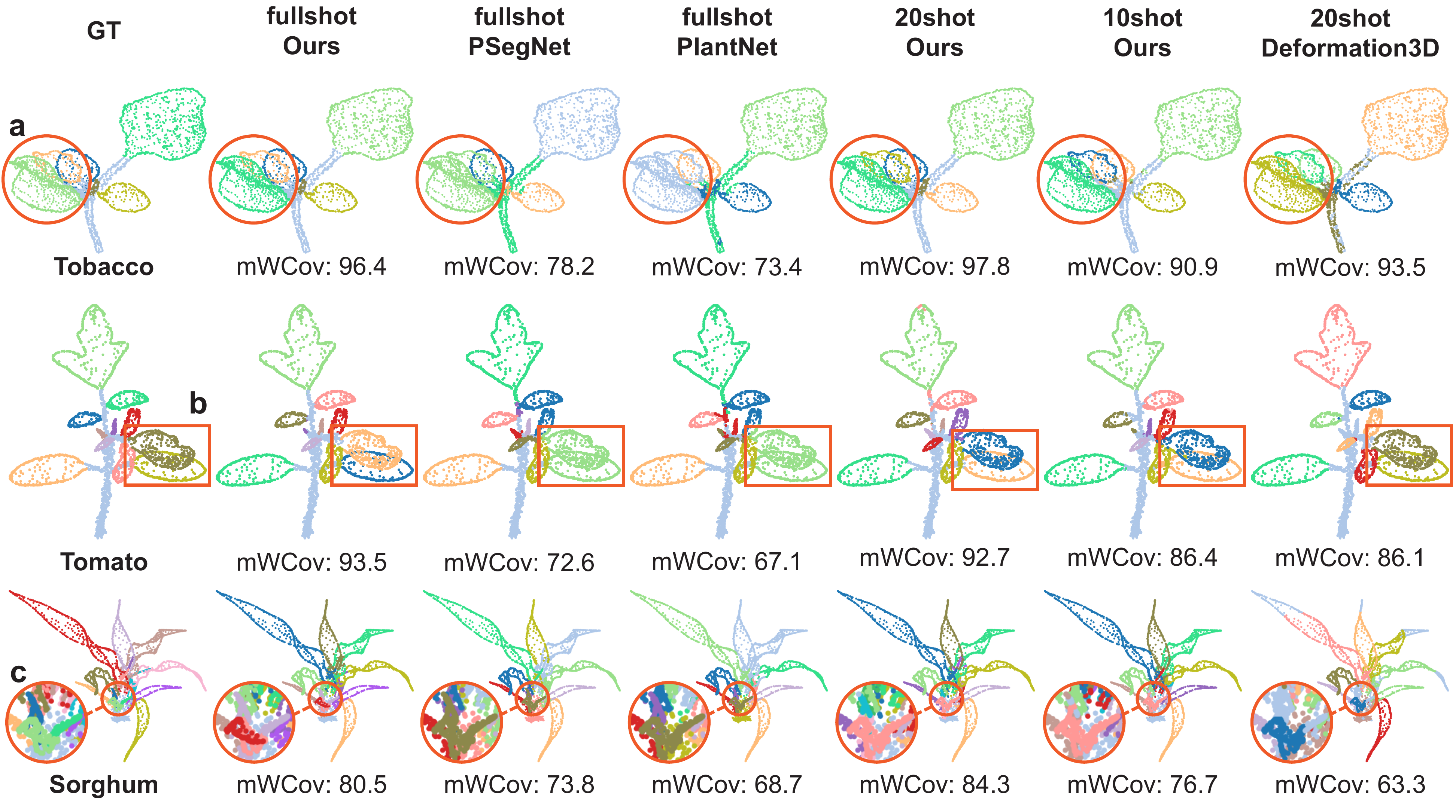}
\caption{Leaf instance segmentation on selected HR3D samples, with tobacco, tomato, and sorghum shown from top to bottom.
Columns from left to right show ground truth (GT), the full-shot predictions of PlantC2USeg, PSegNet, and PlantNet, the 20-shot and 10-shot predictions of PlantC2USeg, and the 20-shot Deformation3D prediction.
Regions~a and~b mark overlapping broad-leaf pairs, while region~c marks three slender leaves at a compact base.
Values denote sample-level mWCov (\%).}\label{fig12}
\end{figure*}

Figure~\ref{fig12} compares full-shot and few-shot leaf instance predictions on the same HR3D samples used for semantic evaluation in Figure~\ref{fig8}.
Overlapping broad leaves exposed repeated under-segmentation by full-shot PSegNet and PlantNet in tobacco region~a and tomato region~b.
Both methods merged each highlighted leaf pair into one instance, whereas full-shot and 20-shot PlantC2USeg separated the two leaves.
At the compact sorghum base in region~c, full-shot PSegNet and 10-shot PlantC2USeg merged all three slender leaves into one instance, while full-shot PlantNet and 20-shot PlantC2USeg partitioned them into two instances.
Full-shot PlantC2USeg separated all three leaves but introduced a spurious partition.
At 20 shots, Deformation3D recovered only one of the three leaves as a distinct instance and omitted the distal tip of another leaf.

Among full-shot methods, PlantC2USeg led every aggregate and species-specific instance metric on HR3D (Table~\ref{table2}).
Relative to PSegNet \citep{li2022psegnet}, the strongest plant-specific joint comparator, its margin was 1.80 percentage points in semantic IoU and reached 7.91 percentage points in mWCov and 13.18 percentage points in mRec.
The multi-species advantage was therefore more pronounced in leaf coverage and ground-truth leaf recovery than in semantic labeling.
In the selected tobacco and tomato examples, PSegNet merged leaf pairs that PlantC2USeg separated, a sample-level relation consistent with the higher mRec of PlantC2USeg (Figure~\ref{fig12}).

The 20-shot PlantC2USeg model ranked second in aggregate mCov, mWCov, and mRec among all evaluated configurations, behind only its full-shot counterpart.
Against Deformation3D \citep{yang2024deformation3d} under the same annotation budget, the mRec advantage reached 16.74 percentage points and mWCov was also higher, while the mPrec advantage was only 1.36 percentage points.

Even with 10 shots, PlantC2USeg exceeded full-shot SGPN \citep{wang2018sgpn} across the coverage and matched-instance measures and exceeded full-shot ASIS \citep{wang2019asis} in mCov, mWCov, and mRec.
Across unequal supervision settings, it also exceeded 20-shot Deformation3D in these three metrics.
Its mWCov reached 79.42\%, compared with 77.45\% for ASIS and 73.91\% for Deformation3D, while mPrec remained lower than for both methods.
Across both limited-label settings, PlantC2USeg retained the advantage in leaf-instance coverage and mRec over ASIS, whereas ASIS retained higher semantic IoU, with the 20-shot mRec margin reaching 20.90 percentage points.

\subsubsection{SYAU-Maize}\label{sec:syau-maize-leaf-instance}
\begin{figure*}[pos=htbp]
\centering
\includegraphics{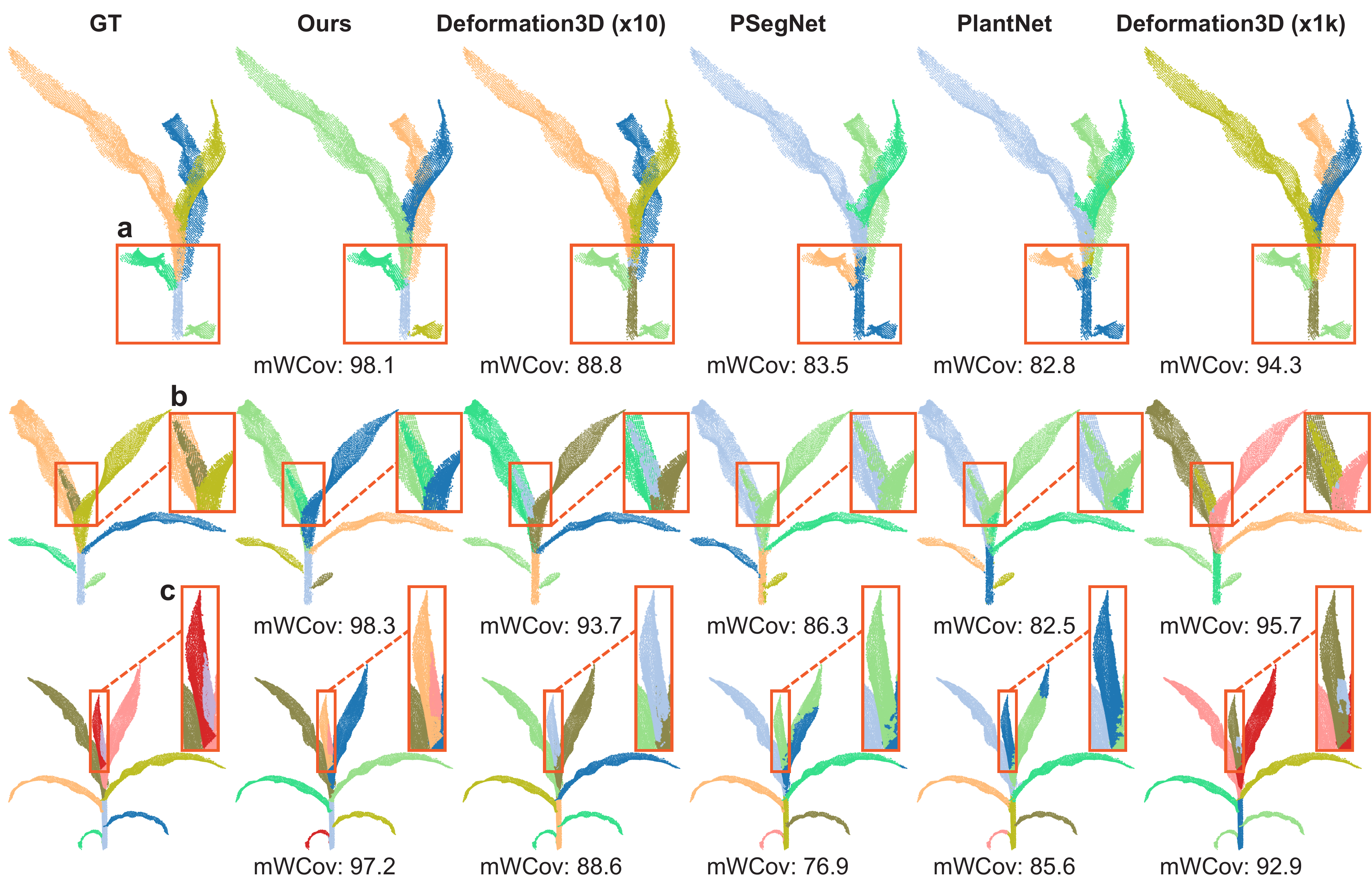}
\caption{Leaf instance segmentation on the same three SYAU-Maize samples and regions as Figure~\ref{fig9} under the common 22-sample annotation budget, with displayed values denoting sample-level mWCov (\%).
Columns from left to right show ground truth (GT), PlantC2USeg, Deformation3D (x10), PSegNet, PlantNet, and Deformation3D (x1k).
The x10 and x1k settings denote tenfold and 1000-fold deformation expansion of the same training set.
Region~a marks a cotyledon near abscission, and regions~b and~c mark closely packed emerging leaves within the central whorl.}\label{fig13}
\end{figure*}

On the same three SYAU-Maize samples and regions used in Figure~\ref{fig9}, leaf instance predictions expose cotyledon assignment near abscission and emerging-leaf separation within the central whorl (Figure~\ref{fig13}).
PlantNet and PSegNet assigned the region~a cotyledon to stem, excluding it from their leaf-instance partitions.
Both Deformation3D variants merged the cotyledon with an adjacent small leaf, whereas PlantC2USeg retained it as a distinct leaf instance.
Regions~b and~c contain rolled emerging leaves closely packed within the central whorl, where limited physical separation and mutual occlusion obscure inter-leaf boundaries, particularly under the tighter enclosure in region~c.
PlantNet and PSegNet merged the central emerging leaf with an enclosing leaf in region~b and, together with Deformation3D (x10), failed to recover the emerging leaf as a distinct, coherent instance under the tighter enclosure in region~c.
PlantC2USeg assigned most of the tightly enclosed emerging leaf to one coherent instance, whereas Deformation3D (x1k) left the corresponding assignments fragmented.
On the complete sample containing region~c, PlantC2USeg reached 97.2\% mWCov, compared with 76.9\% for PSegNet.

Using the same 22 annotated samples, PlantC2USeg exceeded the directly trained joint baselines PlantNet \citep{li2022plantnet} and PSegNet \citep{li2022psegnet} across all four instance metrics on the SYAU-Maize test set (Table~\ref{table3}).
Its margins over both baselines were at least 19.50 percentage points in both coverage measures and mRec.
The semantic and instance leads support the greater effectiveness of the complete PlantC2USeg strategy over both direct joint baselines in the evaluated sparse-label setting.

Within Deformation3D \citep{yang2024deformation3d}, expansion from x10 to x1k raised mCov and mRec, while mWCov rose by 3.38 percentage points as mPrec fell by 3.63 percentage points.
PlantC2USeg was higher than x1k in mCov, mPrec, and mRec, including a 5.26 percentage point mRec margin, whereas x1k retained a 2.67 percentage point lead in mWCov.
Because mCov weights instances equally and mWCov weights them by point count, the aggregate reversal suggests a relative overlap advantage for PlantC2USeg among leaves with fewer points and for x1k among leaves with more points.
Greater deformation expansion therefore favored instance coverage weighted by point count but yielded no uniform cross-task gain, as semantic IoU declined from x10 to x1k.

\subsection{Object part segmentation}
\begin{figure*}[pos=htbp]
\centering
\includegraphics[width=\textwidth]{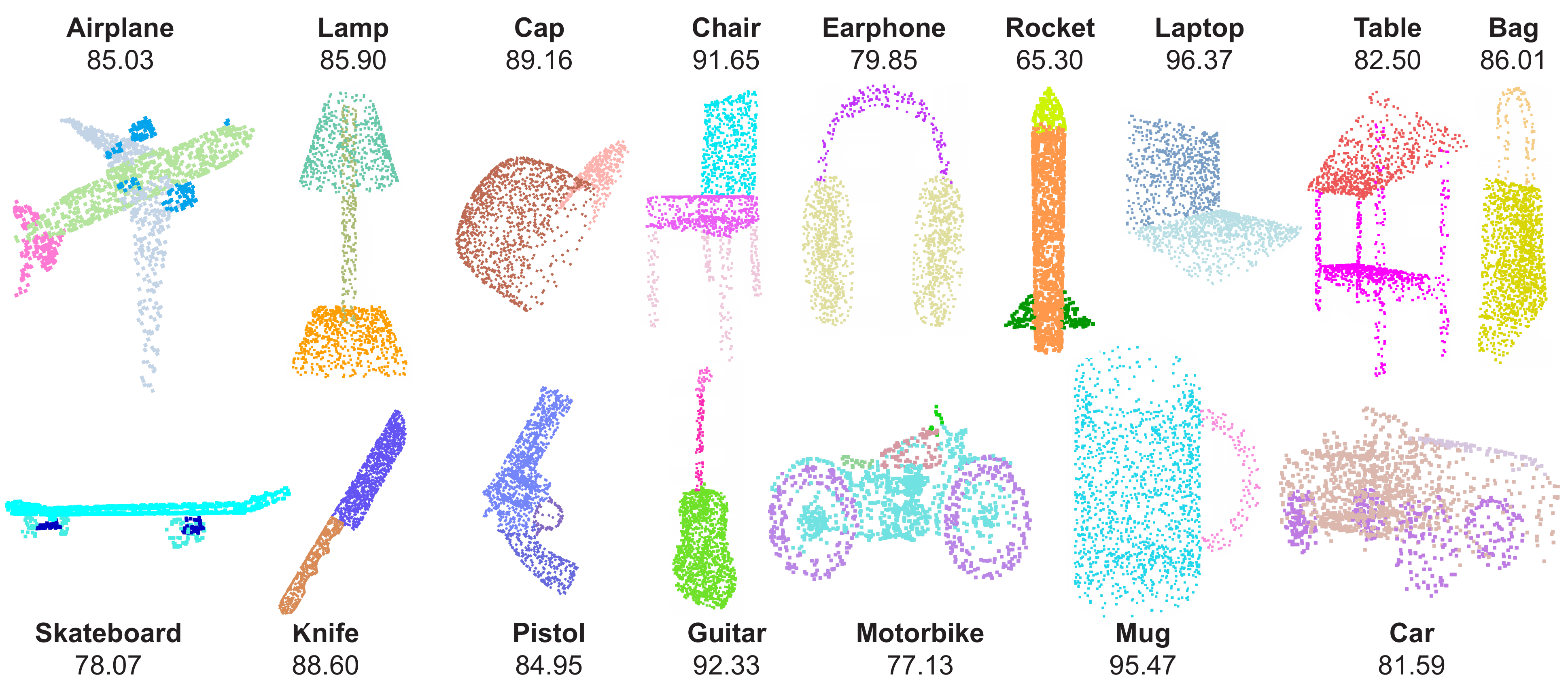}
\caption{Qualitative part segmentation results on the ShapeNet Part dataset, with displayed values denoting category IoU (\%).}\label{fig14}
\end{figure*}

Qualitative results in Figure~\ref{fig14} show that accurate part segmentation is maintained across a wide range of generic object categories.
On ShapeNet Part, PlantC2USeg reached 85.0\% category-averaged mIoU and 86.4\% instance-averaged mIoU, exceeding baselines trained from scratch \citep{qi2017pointnet, qi2017pointnet++, wang2019dgcnn, vaswani2017transformer} and extending the evaluation beyond plant-specific segmentation.
Among single-modal pre-training methods \citep{sauder2019jigsaw3D, wang2021occo, pang2022pointmae, zhang2022pointm2ae, chen2023pointGPT, Zha2024PointFEMAE, zhang2024pointpcpmae, Zhang_2025_PointPQAE}, it ranked first in category-averaged mIoU and second in instance-averaged mIoU, matching the best cross-modal results \citep{dong2023ACT,qi2023recon} for the latter metric.
\begin{table}
\centering
\caption{Quantitative results of part segmentation on ShapeNet Part (\%). Cat. mIoU and Inst. mIoU denote category-averaged and instance-averaged mIoU, respectively.}\label{table4}
\begin{tabular}{lcc}
\hline
Method & Cat. mIoU & Inst. mIoU \\
\hline
\multicolumn{3}{l}{\textit{Training from scratch}}\\
PointNet & 80.4 & 83.7 \\
PointNet++ & 81.9 & 85.1 \\
DGCNN & 82.3 & 85.2 \\
Transformer & 83.4 & 85.1 \\
\cline{1-3}
\multicolumn{3}{l}{\textit{Single-modal pre-training}}\\
Jigsaw3D & 83.0 & 85.3 \\
OcCo & 83.4 & 85.1 \\
Point-MAE & 84.2 & 86.1\\
Point-M2AE & 84.8 & \textbf{86.5}\\
PointGPT &  84.1 & 86.2\\
Point-FEMAE & \underline{84.9} & 86.3\\
PCP-MAE & \underline{84.9} & 86.1\\
Point-PQAE & 84.6 & 86.1\\
PlantC2USeg & \textbf{85.0} & \underline{86.4}\\
\cline{1-3}
\multicolumn{3}{l}{\textit{Cross-modal pre-training}}\\
ACT & 84.7 & 86.1 \\
ReCon & 84.8 & \underline{86.4} \\

\hline
\end{tabular}
\end{table}

\section{Discussion}
\subsection{Ablation study}
The full-shot Soybean3D ablation in Table~\ref{table5} examines cross-scale coherence and reconstruction conditioned on visible context as complementary constraints on plant representation learning for stem--leaf semantic segmentation and leaf instance segmentation.
The Hierarchical Transformer trained from scratch provides an aligned control with the same architecture and downstream objectives but no pre-training.
With standard self-attention and without cross-scale consistency learning, pre-training increased semantic recall and all four instance metrics over scratch initialization but slightly reduced F1 and IoU, exposing uneven transfer between leaf instance segmentation and aggregate stem--leaf discrimination.
\begin{table*}
\centering
\caption{Ablation analysis of each component for stem--leaf and leaf instance segmentation (\%).
C2L denotes cross-scale consistency learning,
IR-Dec denotes information-restricted decoding.}\label{table5}
\begin{tabular}{ccccccccccc}
\hline
Pre-train & C2L & IR-Dec
& \multicolumn{4}{c}{Semantic Segmentation}
& \multicolumn{4}{c}{Instance Segmentation} \\
\cmidrule(lr){4-7} \cmidrule(lr){8-11}
 & & & Prec & Rec & F1 & IoU & mCov & mWCov & mPrec & mRec\\
\hline
           &           &          & \textbf{96.64} & 92.98 & 94.70 & 90.19 & 85.54 & 92.11 & 94.56 & 88.63 \\
\checkmark &           &          & 94.82 & 94.08 & 94.45 & 89.75 & 86.73 & 92.62 & 94.69 & 90.98 \\
\checkmark & \checkmark &          & 95.71 & 94.66 & 95.17 & 91.00 & \underline{88.06} & \underline{93.81} & \underline{95.93} & \underline{92.55} \\
\checkmark &           & \checkmark & 94.96 & \textbf{95.40} & \underline{95.18} & \underline{91.01} & 86.89 & 93.06 & 95.16 & \underline{92.55} \\
\checkmark & \checkmark & \checkmark & \underline{96.40} & \underline{95.01} & \textbf{95.69} & \textbf{91.91} & \textbf{89.94} & \textbf{94.62} & \textbf{96.39} & \textbf{94.12} \\
\hline
\end{tabular}
\end{table*}

By relating corresponding visible regions across spatial scales, cross-scale consistency learning raised IoU by 1.25 percentage points with the standard decoder and by 0.90 percentage points with the information-restricted decoder, with F1 improving in both cases.
The consistent gains support coupling fine-scale organ cues with broader plant architecture during encoder representation learning.
Information-restricted decoding likewise raised F1 and IoU whether cross-scale consistency learning was present or absent.
The information-restricted decoder conditions reconstruction on visible evidence, complementing the cross-scale encoder constraint imposed by cross-scale consistency learning.
The combined design reached 95.69\% F1 and 91.91\% IoU while leading all four instance metrics.
The joint lead across both tasks supports coordinating cross-scale encoder coherence with reconstruction grounded in visible evidence for the representation shared by organ classification and leaf separation.

\subsection{Attention distribution across scales during encoding}
\begin{figure*}[pos=htbp]
\centering
\includegraphics{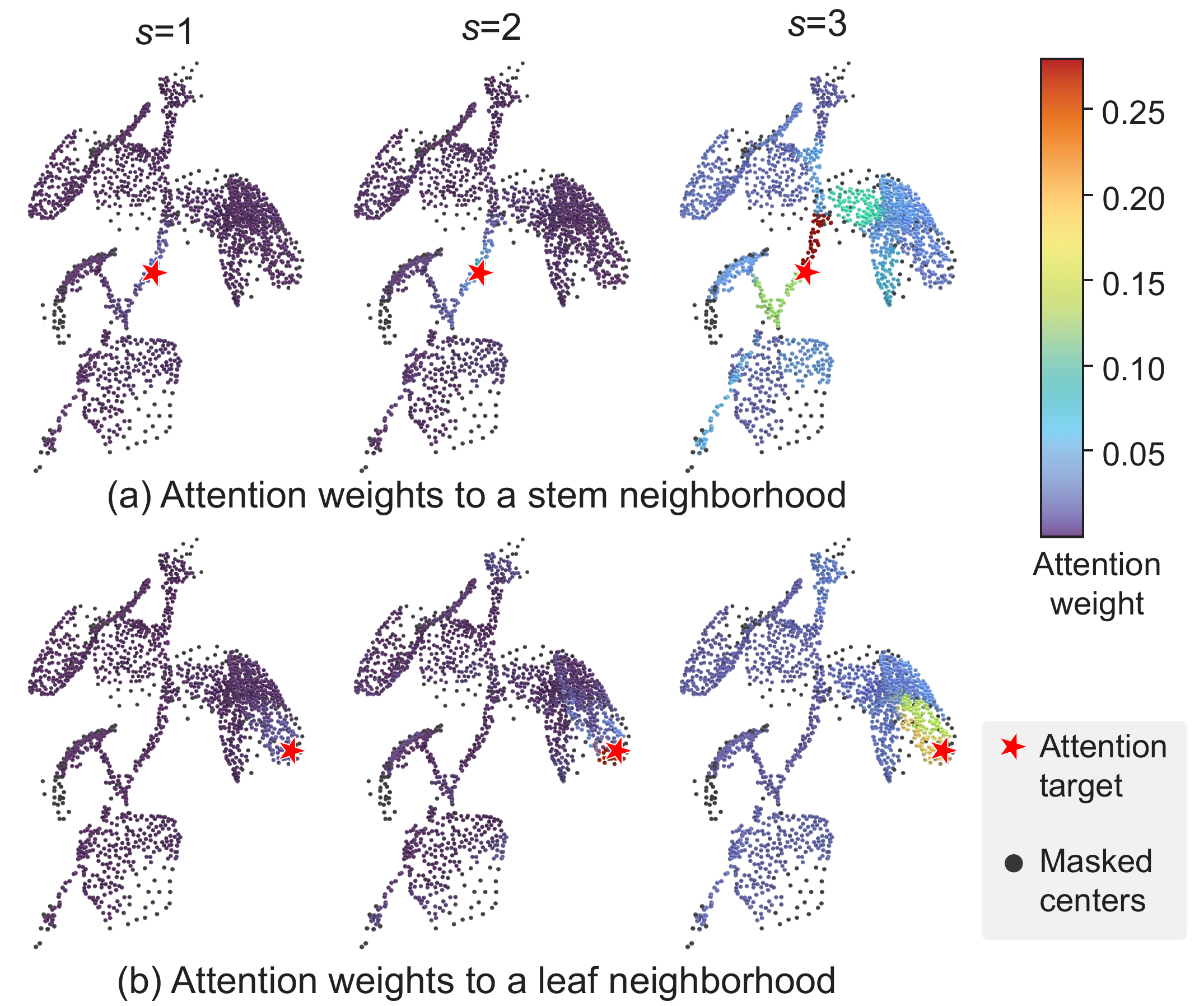}
\caption{Encoder self-attention distributions across hierarchical stages ($s=1,2,3$) for (a) a selected stem target neighborhood and (b) a selected leaf target neighborhood in one representative point cloud.
Red stars mark the attention targets, and black points mark the centers of masked neighborhoods.
Colors use a common displayed scale for the attention weights.}\label{fig15}
\end{figure*}

Encoder attention around selected stem and leaf neighborhoods shows how spatial support develops through hierarchical stages $s=1,2,3$ in one representative point cloud (Figure~\ref{fig15}).
At the early stages, support remains local to the selected neighborhoods.
Attention continues to emphasize the selected stem and leaf neighborhoods across all three stages while reaching a broader portion of the connected plant at $s=3$.
Key counts differ across stages and attention is normalized over keys, so the larger displayed weights at $s=3$ do not by themselves establish stronger or more concentrated attention.
This progression is consistent with the aim of cross-scale consistency learning to relate fine geometric cues to broader plant morphology.

\subsection{Diagnostic comparison of decoder attention allocation}
\begin{figure*}[pos=htbp]
\centering
\includegraphics{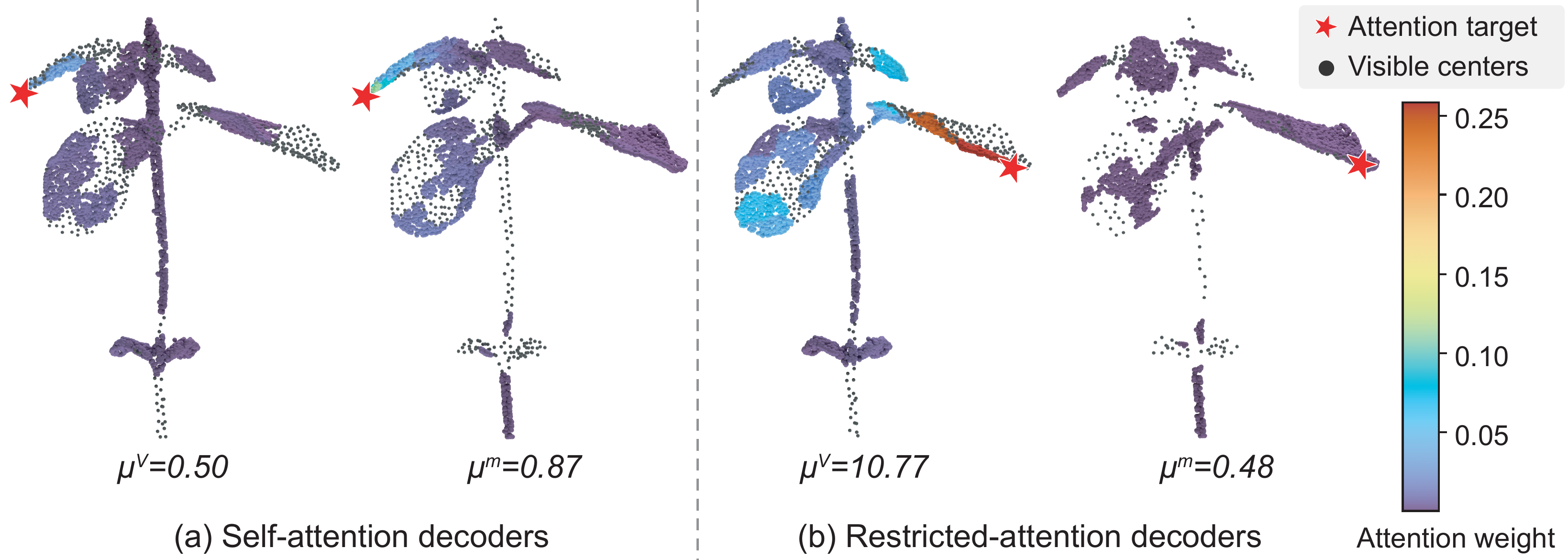}
\caption{Diagnostic comparison of attention allocation for (a) a standard self-attention decoder and (b) a restricted decoder that limits masked-to-masked information exchange in one decoder layer.
The reported $\mu^{v}$ and $\mu^{m}$ are attention masses from masked queries to visible and masked keys, respectively.
Black points mark the centers of visible neighborhoods, and red stars mark the displayed attention targets.
All four maps use a common displayed attention-weight scale.}\label{fig16}
\end{figure*}

Unrestricted decoder self-attention leaves masked tokens able to exchange information during reconstruction.
In the standard diagnostic in Figure~\ref{fig16}, masked queries allocate attention masses of $\mu^{m}=0.87$ to masked keys and $\mu^{v}=0.50$ to visible keys.
This allocation is consistent with masked-token exchange providing an internal reconstruction route.
This route competes with the intended use of evidence encoded from visible neighborhoods during pre-training but belongs to a decoder that is not transferred downstream.
Restricting masked-to-masked exchange in one diagnostic layer reverses the allocation to $\mu^{v}=10.77$ for visible keys and $\mu^{m}=0.48$ for masked keys.
The dominant visible-key mass shows that visible encoder representations contain usable reconstruction evidence.
This allocation reversal motivates the final information-restricted decoder, in which masked queries use projected visible representations as keys and values.

\subsection{Sensitivity of inherited feature-distance criteria}
\begin{figure*}[pos=htbp]
\centering
\includegraphics[width=\textwidth]{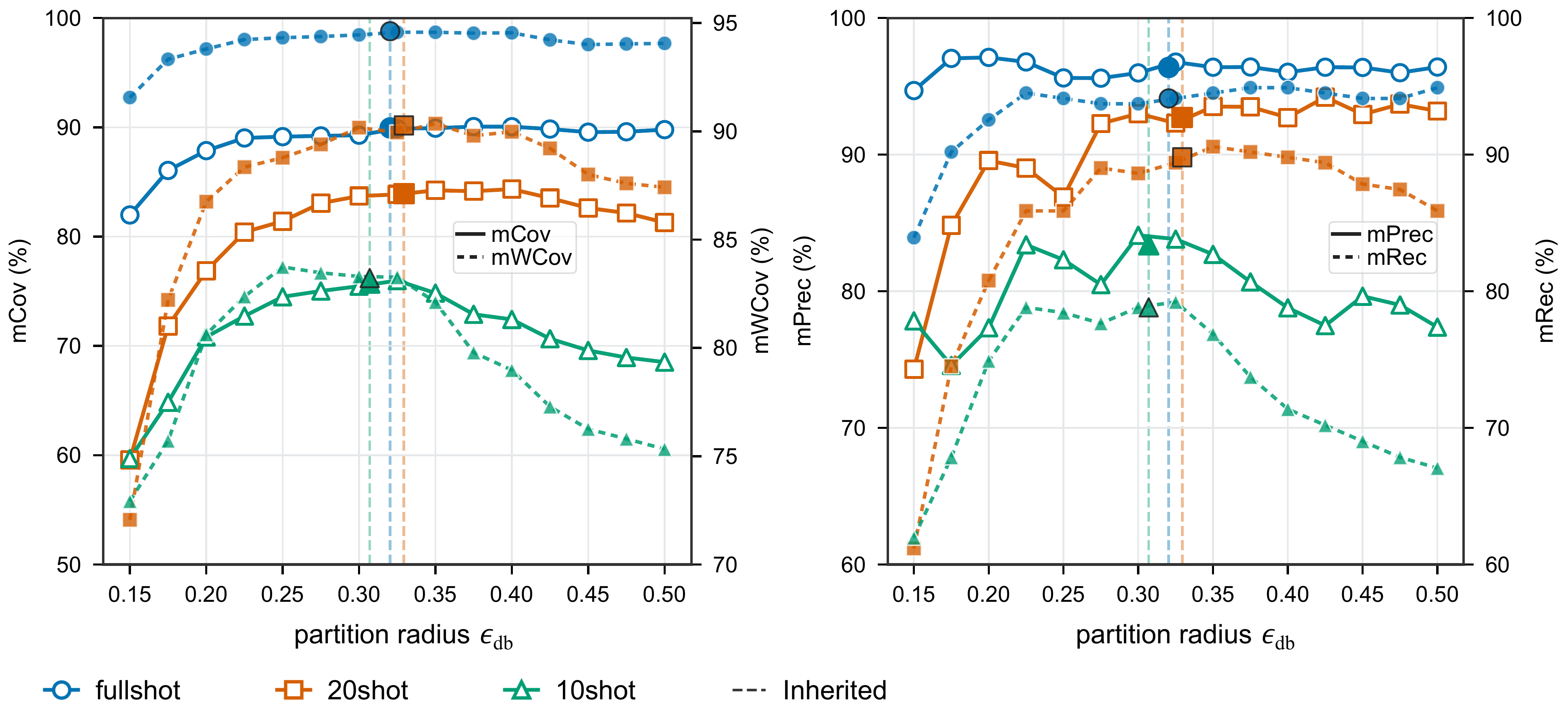}
\caption{Sensitivity of Soybean3D leaf instance segmentation to the feature-space $\ell_1$ partition radius $\epsilon_{\mathrm{db}}$ under full-shot, 20-shot, and 10-shot supervision.
Within each supervision regime, only $\epsilon_{\mathrm{db}}$ varies while $\epsilon_{\mathrm{rec}}$, $\epsilon_{\mathrm{merge}}$, and the checkpoint remain fixed.
Within each regime, enlarged filled markers denote the partition radius inherited from fine-tuning statistics, while the remaining markers show the evaluated sweep values.}
\label{fig17}
\end{figure*}

\begin{figure*}[pos=htbp]
\centering
\includegraphics[width=\textwidth]{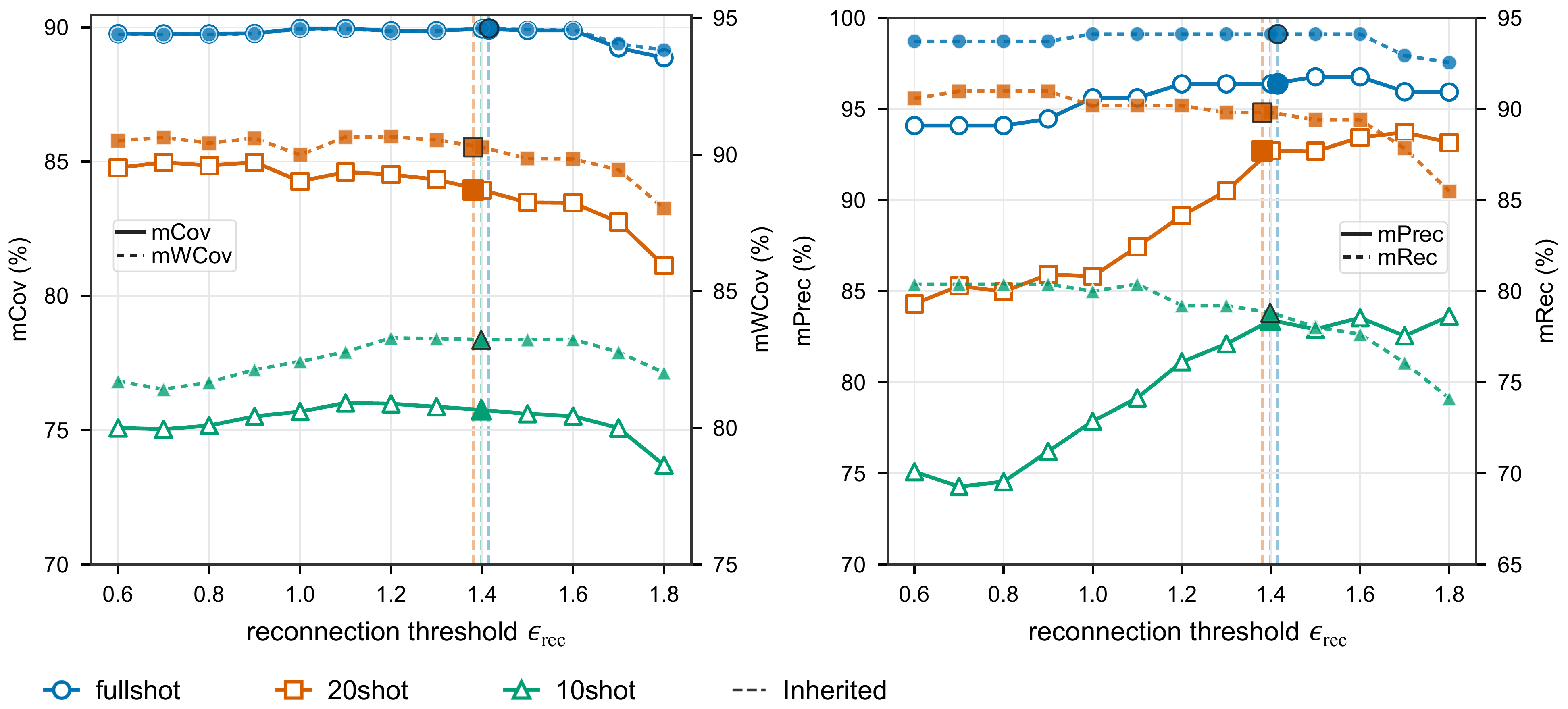}
\caption{Sensitivity of Soybean3D leaf instance segmentation to the feature-centroid $\ell_1$ threshold $\epsilon_{\mathrm{rec}}$ under full-shot, 20-shot, and 10-shot supervision.
This threshold governs whether a later feature partition inherits the nearest existing instance label.
Within each supervision regime, only $\epsilon_{\mathrm{rec}}$ varies while $\epsilon_{\mathrm{db}}$, $\epsilon_{\mathrm{merge}}$, and the checkpoint remain fixed.
Within each regime, enlarged filled markers denote the reconnection threshold inherited from fine-tuning statistics, while the remaining markers show the evaluated sweep values.}
\label{fig18}
\end{figure*}

\begin{figure*}[pos=htbp]
\centering
\includegraphics[width=\textwidth]{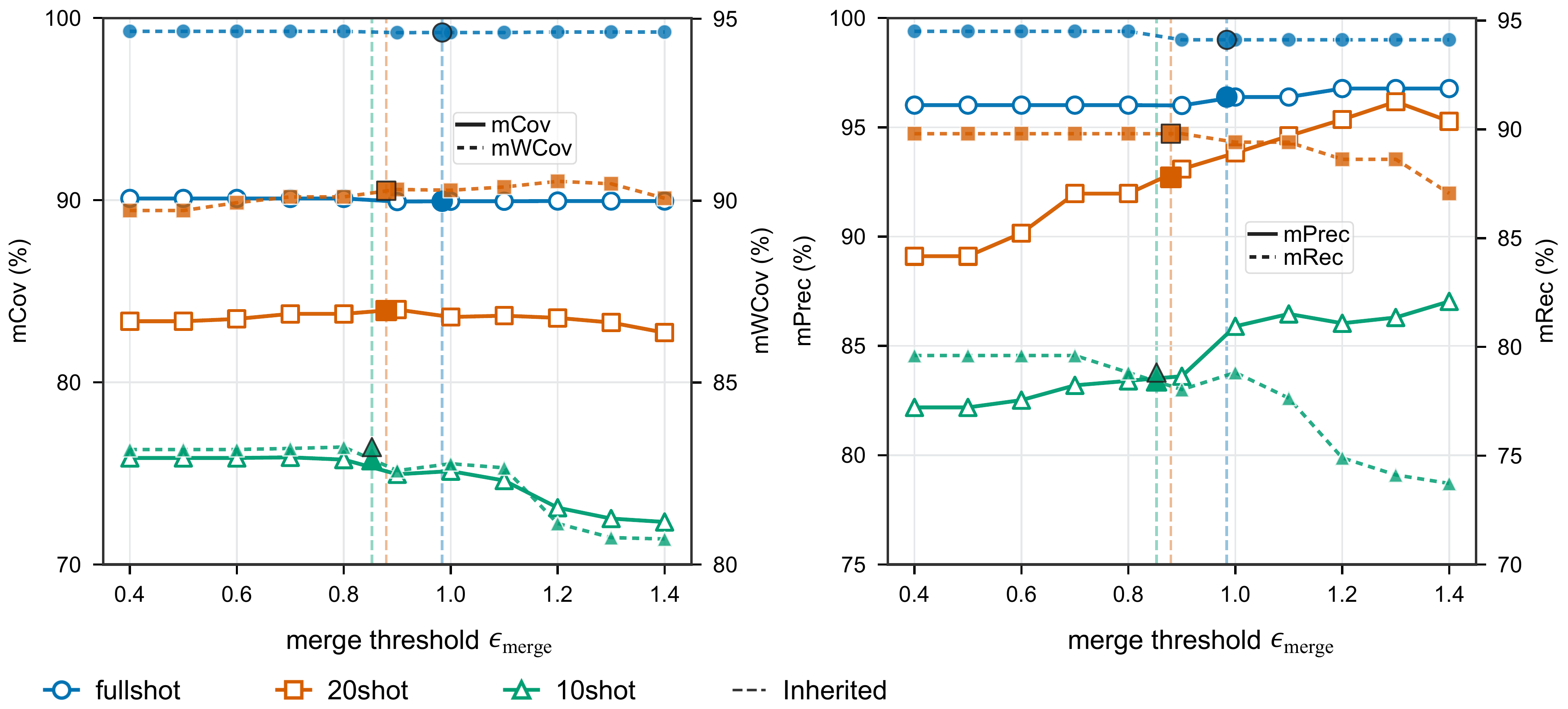}
\caption{Sensitivity of Soybean3D leaf instance segmentation to the feature-centroid $\ell_1$ threshold $\epsilon_{\mathrm{merge}}$ used for final consolidation after residual assignment under full-shot, 20-shot, and 10-shot supervision.
Within each supervision regime, only $\epsilon_{\mathrm{merge}}$ varies while $\epsilon_{\mathrm{db}}$, $\epsilon_{\mathrm{rec}}$, and the checkpoint remain fixed.
Within each regime, enlarged filled markers denote the merge threshold inherited from fine-tuning statistics, while the remaining markers show the evaluated sweep values.}
\label{fig19}
\end{figure*}

Progressive segmentation inherits three feature-distance criteria from fine-tuning statistics, linking the learned embedding space directly to leaf grouping without an independent threshold search after training.
Each criterion was varied separately for Soybean3D leaf instance segmentation under full-shot, 20-shot, and 10-shot supervision while the checkpoint and the other two criteria remained fixed within each regime (Figures~\ref{fig17}--\ref{fig19}).
The partition radius $\epsilon_{\mathrm{db}}$ governs within-step feature-space partitioning, $\epsilon_{\mathrm{rec}}$ governs cross-step inheritance of an existing instance label, and $\epsilon_{\mathrm{merge}}$ governs final consolidation after residual assignment.

The partition radius $\epsilon_{\mathrm{db}}$ had the strongest effect on coverage and mRec, producing the largest mCov, mWCov, and mRec spans among the three criteria in every supervision regime.
Both extremes reduced these outcomes under few-shot supervision (Figure~\ref{fig17}).
At 20 shots, mCov spanned 24.77 percentage points while the inherited value remained 0.41 percentage points below the observed maximum.

The inheritance and consolidation criteria shared a trade-off between mPrec and coverage or mRec, but their response patterns differed.
As $\epsilon_{\mathrm{rec}}$ increased under 20-shot and 10-shot supervision, mPrec generally rose, whereas coverage and mRec peaked at lower thresholds and declined at the upper end (Figure~\ref{fig18}).
Coverage changed little across $\epsilon_{\mathrm{merge}}$ under full-shot supervision.
At 10 shots, the $\epsilon_{\mathrm{merge}}$ value that maximized mPrec raised it by 3.63 percentage points relative to the inherited value but reduced coverage and mRec (Figure~\ref{fig19}).

Across all three criteria and supervision regimes, the inherited values kept mCov and mWCov within 1.04 and 0.51 percentage points of their respective observed maxima while avoiding isolated mPrec maxima associated with lower coverage or mRec.

\subsection{Sensitivity of the initial-radius schedule}
\begin{figure*}[pos=htbp]
\centering
\includegraphics[width=\textwidth]{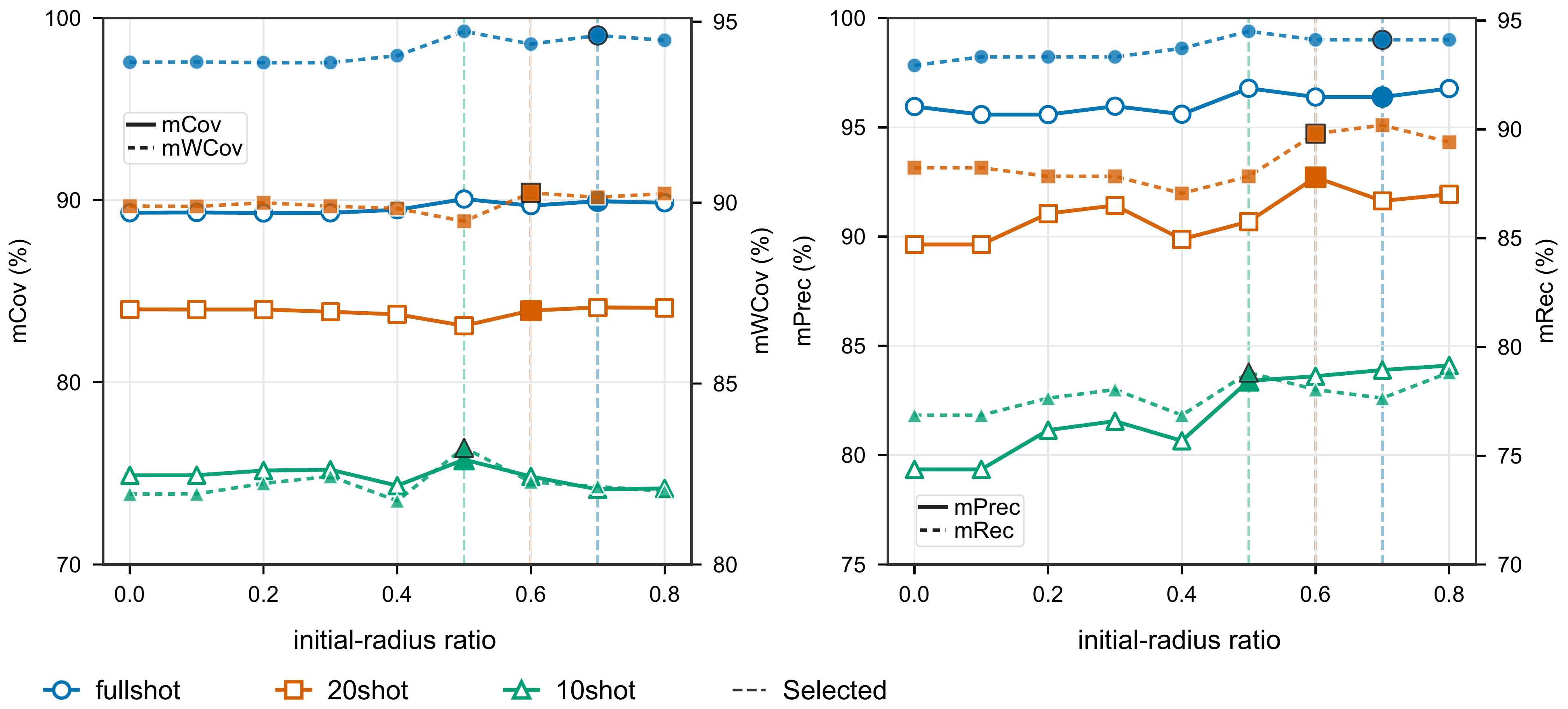}
\caption{Sensitivity of Soybean3D leaf instance segmentation to the initial-radius ratio under full-shot, 20-shot, and 10-shot supervision.
The ratio schedules radial eligibility in shifted planar coordinates before feature-space partitioning.
Only the initial-radius ratio varies from 0.0 to 0.8 while $\epsilon_{\mathrm{db}}$, $\epsilon_{\mathrm{rec}}$, $\epsilon_{\mathrm{merge}}$, and the checkpoint remain fixed within each supervision regime.
Within each regime, enlarged filled markers denote the selected initial-radius ratio, while the remaining markers show the evaluated sweep values.}
\label{fig20}
\end{figure*}

Before feature-space partitioning, the initial-radius ratio schedules radial eligibility in shifted planar coordinates and remains separate from the inherited feature-distance criteria (Figure~\ref{fig20}).
Larger ratios begin with more peripheral candidates, whereas a ratio of zero processes all candidates in one pass.

The selected radial schedules favored broad leaf coverage and matched-instance recovery over marginal gains in precision.
Under full-shot supervision, ratios from 0.5 to 0.8 kept all four instance metrics within 0.41 percentage points of their respective observed maxima and included the selected ratio of 0.7.
At 20 shots, the selected ratio of 0.6 maximized mWCov and mPrec while mCov and mRec remained near their observed maxima.
At 10 shots, increasing the ratio from the selected value of 0.5 to 0.8 raised mPrec by less than 0.70 percentage points and left mRec unchanged.
The same change lowered coverage by at least 1.20 percentage points.

\subsection{Annotation efficiency and practical adaptation}
On Soybean3D, reusable pre-training reduces dependence on target annotations relative to the aligned scratch control across the 20-shot and 10-shot budgets.
At 20 shots, PlantC2USeg pre-training raised semantic IoU by 2.63 percentage points and mWCov, which weights leaf-instance coverage by size, by 8.01 percentage points over random initialization.
At 10 shots, it also maintained higher IoU and mWCov than the same scratch control, extending the aligned benefit across both limited-label budgets.
Under the 22-template SYAU-Maize protocol, PlantNet \citep{li2022plantnet} and PSegNet \citep{li2022psegnet} jointly learn semantic and instance segmentation from the annotated samples, whereas PlantC2USeg applies reusable pre-training before unified adaptation in one shared model.
PlantC2USeg exceeded both fully supervised baselines by about 9 percentage points in IoU and about 20 percentage points in mWCov, supporting reusable representation transfer under this annotation budget.

The comparison with target-specific deformation spans Soybean3D, HR3D, and SYAU-Maize, which together cover five crop species and use MVS reconstruction, blue-laser scanning, and high-precision 3D scanning, respectively.
Deformation3D \citep{yang2024deformation3d} generates target-specific training clouds and trains separate PointNet++ semantic and HAIS instance models, whereas PlantC2USeg reuses a pre-trained representation before target fine-tuning.
Under matched original annotation budgets, PlantC2USeg exceeded Deformation3D x10 by 1.56--15.38 percentage points in semantic IoU and attained higher mRec on all three datasets.
Across the three datasets, these consistent IoU and mRec gains support representation reuse over repeated target-specific data generation within the evaluated plant distributions.
Target-specific generation also repeats data preparation for each template, with x1k preparation projected at 2.31 h for one 590,574-point Soybean3D template.

Moving Deformation3D from x10 to x1k increased the number of generated clouds 100-fold, raised mWCov by 3.38 percentage points, and lowered IoU by 0.60 percentage points.
Among the five methods evaluated under this protocol, Deformation3D x1k attained the highest mWCov, while PlantC2USeg remained higher than x1k in semantic IoU and the other three instance metrics.
In this comparison, larger expansion improves size-weighted instance coverage at the expense of semantic overlap.

PlantC2USeg fine-tunes one shared model to produce semantic predictions, feature embeddings, and coordinate offsets, while deriving the three criteria used in progressive segmentation directly from fine-tuning statistics.
This unified adaptation contrasts with the separate downstream networks of Eff-3DPSeg \citep{luo2023eff3dpseg}, the sequential adaptation of Soybean-PCMAE \citep{tian2025soybeanpcmae}, and the separate semantic and instance models of Deformation3D \citep{yang2024deformation3d}.

Table~\ref{table6} reports network-core forward and backward computation for one Soybean3D sample after the native input processing of each method.
PlantC2USeg recorded the shortest time among the five evaluated methods, 38.60 ms compared with 69.60 ms for the next-fastest HAIS.
The timed PlantC2USeg model covers semantic and instance segmentation in one shared network, whereas the HAIS measurement covers only the Deformation3D instance network.
The measured speed advantage therefore supports the practical efficiency of unified adaptation within the reported network-core scope.
\begin{table*}
\centering
\caption{Controlled training efficiency for one Soybean3D sample.
For parameters, FLOPs, both timing measures, and peak memory, lower values are preferable.
All methods were profiled on an NVIDIA GeForce RTX 3090 at batch size 1 after five warmup iterations and over 20 timed iterations.
All methods use the same Soybean3D sample, which contains 590,574 points before method-specific input processing.
PlantNet, PSegNet, and PlantC2USeg process 4,096 sampled points.
Eff-3DPSeg voxelizes the complete sample, whereas Deformation3D samples 20,480 points before voxelization.
Forward and backward time excludes data loading, augmentation, voxelization, optimizer updates, logging, validation, and checkpointing.
Methods are ordered by the combined forward and backward time.}\label{table6}
\small
\setlength{\tabcolsep}{4pt}
\begin{tabular}{lccccccc}
\hline
Method
& \shortstack{Input\\points}
& \shortstack{Active\\voxels}
& \shortstack{Parameters\\(M)}
& \shortstack{FLOPs\\(G)}
& \shortstack{Forward time\\(ms)}
& \shortstack{Forward + backward\\(ms)}
& \shortstack{Peak memory\\(MiB)} \\
\hline
PlantC2USeg & 4,096 & -- & 13.61 & \underline{9.01} & \textbf{18.31} & \textbf{38.60} & \underline{629.6} \\
Deformation3D (HAIS) & 20,480 & 1,861 & 30.84 & \textbf{0.93} & \underline{33.58} & \underline{69.60} & \textbf{244.9} \\
PlantNet & 4,096 & -- & \underline{4.44} & 12.15 & 149.79 & 213.83 & 755.2 \\
PSegNet & 4,096 & -- & \textbf{3.34} & 10.80 & 157.01 & 225.15 & 788.2 \\
Eff-3DPSeg & 590,574 & 231,971 & 37.86 & 490.50 & 107.47 & 503.60 & 2,962.4 \\
\hline
\end{tabular}
\end{table*}

\section{Conclusion}
Reusable pre-training in PlantC2USeg reduces dependence on target annotations while supporting stem--leaf semantic segmentation and leaf instance segmentation through a shared adaptation model.
The full-shot ablation shows that cross-scale consistency learning and reconstruction conditioned on visible evidence impose complementary constraints on the representation adapted jointly for both tasks.
Against the aligned Hierarchical Transformer trained from scratch on full-shot Soybean3D, complete pre-training increased IoU by 1.72 percentage points and mWCov by 2.51 percentage points.
Complete pre-training retained higher IoU and mWCov than random initialization as Soybean3D supervision decreased.
Across Soybean3D, HR3D, and\linebreak SYAU-Maize, higher IoU and mRec than Deformation3D x10 support reusing the pre-trained representation under different target distributions.

The limited-label regimes across the three plant datasets show that PlantC2USeg can produce the stem--leaf labels and individual-leaf partitions required for later organ-level trait extraction from small annotated collections.
Extending PlantC2USeg beyond static point clouds of individual plants requires evaluation in complex multi-plant field scenes.
Inter-plant overlap and occlusion, together with background vegetation and variations in reconstruction completeness and point density, require segmentation to distinguish neighboring plants and associate each stem and leaf with the correct plant.
Temporal phenotyping then requires correspondence between the same organs across repeated 3D observations.
Stable organ correspondence would enable longitudinal measurement of organ-specific growth trajectories across developmental stages.

\appendix

\printcredits

\section*{Declaration of competing interest}
The authors declare that they have no known competing financial interests or personal relationships that could have appeared to influence the work reported in this paper.
\section*{Acknowledgments}
This research is supported by funding from the FRQNT \& MAPAQ Partnership Research Program-Sustainable Agriculture (Grant No. 259806), and the Natural Sciences and Engineering Research Council of Canada (NSERC) Discovery Grants Program (Grant no. G256643).

\section*{Data availability}
Data will be made available on request.

\bibliographystyle{cas-model2-names}

\bibliography{cas-refs}

\end{document}